\documentclass[twoside,11pt]{article}

\usepackage{comment}
\usepackage[utf8]{inputenc} 
\usepackage[T1]{fontenc}    
\usepackage[T1]{fontenc}    
\usepackage{url}            
\usepackage{booktabs}       
\usepackage{amsfonts}       
\usepackage{nicefrac}       
\usepackage{microtype}      
\usepackage{xcolor}         
\usepackage{graphicx}
\usepackage{subcaption}
\usepackage{multirow}
\usepackage[abbrvbib, preprint]{jmlr2e}

\usepackage{hyperref}       
\usepackage{xpatch}
\usepackage{amsmath}
\usepackage[all]{nowidow}
\usepackage{lastpage}

\usepackage[backend=biber]{biblatex}
\usepackage{pict2e}
\usepackage{tikz}
\usetikzlibrary{positioning}
\usepackage{sidecap}
\usetikzlibrary{decorations.pathreplacing}

\graphicspath{
    {images/}
}
\newcommand{\argmaxD}{\arg\!\max}
\usepackage{array}

\jmlrheading{xx}{2024}{1-\pageref{LastPage}}{03/24; Revised xx/yy}{xx/yy}{21-0000}{E. Nyiri and O. Gibaru}

\ShortHeadings{R-LRP XML}{E. Nyiri and O. Gibaru}
\firstpageno{1}
 
\begin{document}

\title{Relative Layer-Wise Relevance Propagation: a more Robust Neural Networks eXplaination}
\author{
   \name Eric Nyiri \email eric.nyiri@ensam.eu \\
   \addr Arts et Métiers Institute of Technology\\
   LISPEN Lille campus, 8 bd Louis XIV 59800 Lille, France
   \AND
   \name Olivier Gibaru\thanks{corresponding author - www.oliviergibaru.org} 
   \email olivier.gibaru@ensam.eu\\
   \addr Arts et Métiers Institute of Technology\\
   LISPEN Lille campus, 8 bd Louis XIV 59800 Lille, France\\
   partial secondment at Buawei
}

\editor{xxx}

\maketitle

\begin{abstract}
   Machine learning methods are solving very successfully a plethora of tasks, but they have the disadvantage of not providing any information about their decision. Consequently, estimating the reasoning of the system provides additional information. For this, Layer-Wise Relevance Propagation (LRP) is one of the methods in eXplainable Machine Learning (XML). Its purpose is to provide contributions of any neural network’s output in the domain of its input. The main drawback of current methods is mainly due to division by small values. To overcome this problem, we provide a new definition called Relative LRP where the classical conservation law is satisfied up to a multiplicative factor but without divisions by small values except for Resnet skip connection.
   In this article, we will focus on image classification. This allows us to visualize the contributions of a pixel to the predictions of a multi-layer neural network. Pixel contributions provide a focus to further analysis on regions of potential interest. R-LRP can be applied for any dense, CNN or residual neural networks. Moreover, R-LRP doesn't need any hyperparameters to tune contrary to other LRP methods. We then compare the R-LRP method on different datasets with simple CNN, VGG16, VGG19 and Resnet50 networks.
\end{abstract}


\section{LRP Introduction}
\label{LRP}
XML methods are especially useful in safety-critical domains like finance, vision control, robotics \cite{Guerin2018a} \cite{Guerin2018b} \cite{Simao2019a} \cite{Simao2019b} \cite{Guerin2021} or gesture assistance \cite{Simao2017} \cite{Simao2019c} \cite{Simao2019d}  where engineers (resp. practitioners for medicine)
must know exactly what the network is paying attention to.
Other use-cases are the diagnosis of network behavior/misbehavior and improving the network architecture. The umbrella term eXplainable Machine Learning or explainable AI can be used for many different techniques \cite{Holzinger2022}.

LIME \cite{Ribeiro2016} and RISE \cite{Petsiuk2018} methods consider networks as black-boxes. They produce relevant regions in saliency maps in order to explain the decision of a network. As explained in \cite{Covert2021}, these techniques estimate importance by probing the model with randomly masked versions of the input image. They then obtained the corresponding outputs, which are used to produce the map. The advantage is that they don't need to know the structure of networks. The main drawback is that they cannot provide fine-grained pixel analysis in the decision-making process. This is mostly due to the fact that the masks cannot cover the area of influence of each input/pixel independently.

To overcome this problem, interpretability is needed. Two closely related branches of interpretability methods are propagation-based and gradient-based approaches. They either propagate the algorithm’s decision back through the model or make use of the sensitivity information provided by gradients of the loss. Prominent representatives are Deconvolution Networks, Guided Backpropagation, Grad-CAM, Integrated Gradients, and Layer-Wise Relevance Propagation (LRP). In this article, we will mainly address LRP methodology \cite{Bach2015}, \cite{Montavon2017}, \cite{Montavon2019}, \cite{Lapuschkin2019}, \cite{Kohlbrenner2020}.
Some methods are very closely related to LRP. In \cite{Fergus2014}, the authors visualized neuron activation with a deconvolutional network. In \cite{Simonyan2015}, saliency maps are computed based on Taylor decomposition. \cite{Springenberg2015} addressed Guided Backpropagation soon afterward. \cite{Zhang2018} introduced excitation backprop which casts the backpropagation as a probabilistic process. \cite{Selvaraju2017} defined Grad-Cam for CNNs and \cite{Sundararajan2017} presented the Integrated Gradient method. Gradient-based approaches are problematic because activation functions such as Rectified Linear Units (ReLUs) have a gradient of zero when they are not firing, and yet a ReLU that does not fire can still carry information. Similarly, $sigmoid$ or $tanh$ activations are popular choices for the activation functions of gates in memory units of recurrent neural networks such as GRUs \cite{Chung2014} and LSTMs \cite{Hochreiter1997}. These activations have a near-zero gradient at high or low inputs, even though such inputs can be very significant.

Intuitively, LRP uses the network weights and the neural activations created by the forward-pass to propagate the output back through the network up until the input layer. There, pixels that really contributed to the output can be visualized. The magnitude
of each pixel's contribution or intermediate neuron are called “relevance” values. \cite{Lapuschkin2017} have used the LRP-method to investigate networks predicting gender and age from image data. \cite{Thomas2019} have applied the technique to a large set of MRI neuroimaging data, explaining brain states from 3D data. \cite{Arras2017} have applied LRP to text, to investigate how a neural network classifies text as belonging to one topic or another. \cite{Horst2019} analyzed human gait patterns with LRP. \cite{Srinivasan2017} used LRP to find out exactly what parts of a video are used by a classifier for human action recognition. \cite{Montavon2019} have applied LRP to Atari games and image classification.

 In this article, we define a new LRP method called R-LRP which is more robust and gives better pixels' contributions than other current LRP methods.
 To visualize pixels' contributions, authors usually use heatmaps, but these representations are subject to interpretation and depend on heatmaps settings. We prefer to visualize contributions in terms of percentage of most relevant inputs issued from each RGB channel separately or pixels in images. R-LRP source code is available at \href{https://gitlab.ensam.eu/NYIRI/r-lrp}{https://gitlab.ensam.eu/NYIRI/r-lrp}

\subsection{DAGs for feed-forward networks}

The typical setting for supervised learning is to find an approximation of an unknown function $f$, given a training data set of its point values: $\{x^{(i)},f(x^{(i)})\}_{i=1,...,m}$ where $x^{(i)} \in \mathbb{R}^d$ and $f(x^{(i)}) \in \mathbb{R}^p$. For example, in classification problems for images, $d$ is the number of pixels $\times$ the number of channels in the images, typically somewhere in the range of $10^3$ to $10^6$, and for videos it is even higher.

The learning problem is then to numerically produce from this data a function $\hat{f}$ that is in some sense a good predictor of $f$ on new unseen draws $x \in \mathbb{R} ^d$. The data fitting task is formulated in a stochastic setting, where one assumes that the data comes from random draws of the data sites $x^{(i)}$ with respect to a probability distribution, and the values $f(x^{(i)})$ are noisy observations of some unknown function $f$.

In the following, we will consider "good predictor" based on feed-forward neural network. They can be associated to a {\bf directed acyclic graph} (or DAG) called the {\it architecture} of the neural network \cite{DeVore2021}.

\begin{definition}[Feed-forward neural network]
\label{def:1}
A feed-forward neural network $\mathcal{N}$ is defined by a {\bf directed acyclic graph} $\mathcal{G}=(\mathcal{V},\mathcal{E})$ where
    \begin{enumerate}
        \item $\mathcal{V}$ is a finite set of vertices and $\mathcal{E}$ is a finite set of directed edges
        \item $\mathcal{I}$ is the subset of {\bf 'source-input'} vertices of $\mathcal{V}$ which do not have any incoming edges from $\mathcal{E}$
        \item $\mathcal{O}$ is the subset of {\bf 'sink-output'} vertices of $\mathcal{V}$ which do not have any outgoing edges from $\mathcal{E}$
        \item for all directed edge $e \in \mathcal{E}$, we associate a {\bf weight} $w_e \in \mathbb{R}$
        \item for all vertex $v \in  \mathcal{V}$
            \begin{enumerate}
                \item $v$ must belong to at least one edge $e \in \mathcal{E}$
                \item we associate to $v$ a {\bf neuron} output
                $$x_v :=
                \begin{cases}
                    \sigma_{v} \left( b_v + \sum\limits_{v' \in Prev(v)} w_e.x_{v'} \right) & \text{if } v \in \mathcal{V}\setminus \mathcal{I}  \\
                    \text{the given input value for vertex } v & \text{ else} \\
                \end{cases}$$
                
                with
                \begin{itemize}
                    \item $\sigma_v:\mathbb{R}\to\mathbb{R}$ being an {\bf activation function} and $b_v\in \mathbb{R}$ being a {\bf bias},
                    \item $Prev(v)$ being the subset of all the previous vertices of any edge in $\mathcal{E}$ terminating at $v$
                \end{itemize}
            \end{enumerate}
     \end{enumerate}

The vertices in $\mathcal{H} := \mathcal{V}\setminus \{\mathcal{I},\mathcal{O}\}$ are called {\bf hidden}.
\end{definition}

As we can see, each vertex is associated to one neuron. The DAGs admit topological sorting and no cycles \cite{Graphs92}. This allows us to deduce relationships in a logical order. Consequently, for any given feed-forward neural network, we can compute the list of all paths from one input vertex of $\mathcal{I}$ to any output vertex of $\mathcal{O}$ and conversely. Moreover, for any $e \in \mathcal{E}$ and $ v \in \mathcal{V} \setminus \mathcal{I}$, the parameters $\{w_e,b_v\}$ are then trainable.

\begin{figure}[t]
	\centering
    \begin{tikzpicture}[
    init/.style={
    shorten >=1pt,
    draw,
    circle,
    inner sep=2pt,
    font=\Huge,
    join = by -latex
    },
    squa/.style={
    draw,
    inner sep=2pt,
    font=\Large,
    join = by -latex
    }  
    ]
        \tikzstyle{unit}=[draw,shape=circle,minimum size=0.75cm]
 
        \node[unit,fill=green!80!](x0) at (-3.5,3.5){$x_0$};
        \node[unit,fill=green!50!](x1) at (-3.5,1.75){$x_1$};
        \node[unit,fill=green!30!](x2) at (-3.5,0){$x_2$};
        
        \node[unit,fill=blue!80](x01) at (0,4.5){$\displaystyle\Sigma$};
        \node[unit,fill=blue!80](x02) at (0,3.5){$\displaystyle\Sigma$};
        \node[unit,fill=blue!80](ex01) at (1,4.5){\tiny $e^x$};
        \node[unit,fill=blue!80](ex02) at (1,3.5){\tiny $e^x$};

        \node[unit,fill=blue!50](x10) at (0,2.25){$\displaystyle\Sigma$};
        \node[unit,fill=blue!50](x12) at (0,1.25){$\displaystyle\Sigma$};
        \node[unit,fill=blue!50](ex10) at (1,2.2){\tiny $e^x$};
        \node[unit,fill=blue!50](ex12) at (1,1.25){\tiny $e^x$};

        \node[unit,fill=blue!30](x20) at (0,0){$\displaystyle\Sigma$};
        \node[unit,fill=blue!30](x21) at (0,-1){$\displaystyle\Sigma$};
        \node[unit,fill=blue!30](ex20) at (1,0){\tiny $e^x$};
        \node[unit,fill=blue!30](ex21) at (1,-1){\tiny $e^x$};

        \node[unit,fill=brown!80](sx0) at (3.5,3.5){$\displaystyle\Sigma$};
        \node[unit,fill=brown!50](sx1) at (3.5,1.75){$\displaystyle\Sigma$};
        \node[unit,fill=brown!30](sx2) at (3.5,0){$\displaystyle\Sigma$};
        \node[unit,fill=brown!80](ix0) at (4.5,3.5){\tiny $\frac{1}{x}$};
        \node[unit,fill=brown!50](ix1) at (4.5,1.75){\tiny $\frac{1}{x}$};
        \node[unit,fill=brown!30](ix2) at (4.5,0){\tiny $\frac{1}{x}$};

        \node[ right=of ix0] (outx0) {$\frac{e^{x0}}{e^{x0}+e^{x1}+e^{x2}}$};
        \node[ right=of ix1] (outx1) {$\frac{e^{x1}}{e^{x0}+e^{x1}+e^{x2}}$};
        \node[ right=of ix2] (outx2) {$\frac{e^{x2}}{e^{x0}+e^{x1}+e^{x2}}$};
        
        \draw[->,color=red!80] (x0) -- (x01) node[pos=.8,above]{$-1$};
        \draw[->,color=red!80] (x0) -- (x02) node[pos=.8,above]{$-1$};
        \draw[->,color=green!50] (x1) -- (x01) node[pos=.9,above]{$1$};
        \draw[->,color=green!30] (x2) -- (x02) node[pos=.9,above]{$1$};
        
        \draw[->,color=red!50] (x1) -- (x10) node[pos=.8,above]{$-1$};
        \draw[->,color=red!50] (x1) -- (x12) node[pos=.8,above]{$-1$};
        \draw[->,color=green!80] (x0) -- (x10) node[pos=.9,above]{$1$};
        \draw[->,color=green!20] (x2) -- (x12) node[pos=.9,above]{$1$};

        \draw[->,color=red!30] (x2) -- (x20) node[pos=.8,above]{$-1$};
        \draw[->,color=red!30] (x2) -- (x21) node[pos=.8,above]{$-1$};
        \draw[->,color=green!50] (x1) -- (x21) node[pos=.9,above]{$1$};
        \draw[->,color=green!80] (x0) -- (x20) node[pos=.9,above]{$1$};
        
        \draw[->,color=blue!80] (x01) -- (ex01) node[pos=.5,above]{};
        \draw[->,color=blue!80] (x02) -- (ex02) node[pos=.5,above]{};
        \draw[->,color=blue!80] (2.5,4.5) -- (sx0) node[pos=.5,above]{$1$};
        \draw[->,color=blue!80] (ex01) -- (sx0) node[pos=.5,above]{$1$};
        \draw[->,color=blue!80] (ex02) -- (sx0) node[pos=.5,above]{$1$};

        \draw[->,color=blue!50] (x10) -- (ex10) node[pos=.5,above]{};
        \draw[->,color=blue!50] (x12) -- (ex12) node[pos=.5,above]{};
        \draw[->,color=blue!50] (2.5,2.5) -- (sx1) node[pos=.5,above]{$1$};
        \draw[->,color=blue!50] (ex10) -- (sx1) node[pos=.5,above]{$1$};
        \draw[->,color=blue!50] (ex12) -- (sx1) node[pos=.5,above]{$1$};

        \draw[->,color=blue!30] (x20) -- (ex20) node[pos=.5,above]{};
        \draw[->,color=blue!30] (x21) -- (ex21) node[pos=.5,above]{};
        \draw[->,color=blue!30] (2.5,0.5) -- (sx2) node[pos=.5,above]{$1$};
        \draw[->,color=blue!30] (ex20) -- (sx2) node[pos=.5,above]{$1$};
        \draw[->,color=blue!30] (ex21) -- (sx2) node[pos=.5,above]{$1$};

        \draw[->,color=brown!80] (sx0) -- (ix0) node[pos=.5,above]{};
        \draw[->,color=brown!50] (sx1) -- (ix1) node[pos=.5,above]{};
        \draw[->,color=brown!30] (sx2) -- (ix2) node[pos=.5,above]{};
        
        \draw[->,color=brown!80] (ix0) -- (outx0) node[pos=.5,above]{};
        \draw[->,color=brown!50] (ix1) -- (outx1) node[pos=.5,above]{};
        \draw[->,color=brown!30] (ix2) -- (outx2) node[pos=.5,above]{};
    
    \end{tikzpicture}
	\caption[Sofmax.]{Softmax implementation example in 1D.}
	\label{fig:softmax}
\end{figure}
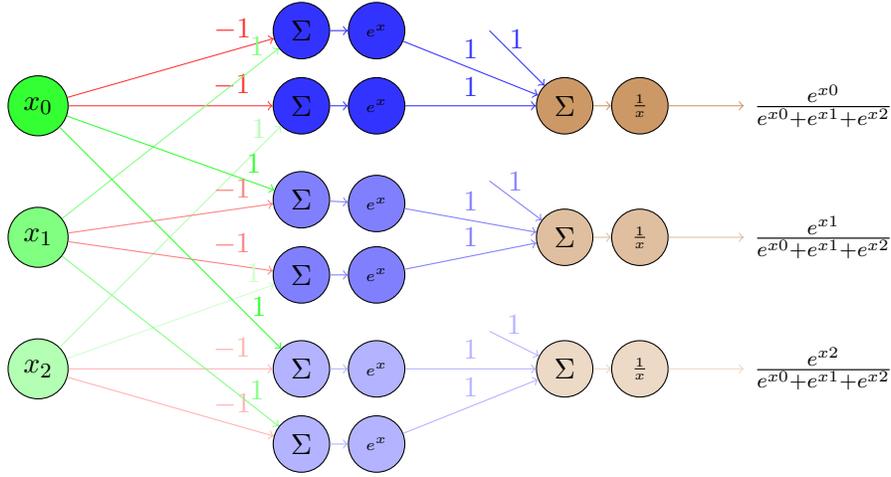

It's a very general definition of feed-forward neural networks. We will see that it encompasses virtually all network architecture used in practice: fully connected, convolutional networks, residual networks, Unet, U2net, Transformers. In order to respect the Definition notations for the Softmax function implementation,  we have to add intermediate layers of neurons with the appropriate activation functions, weights and biases (see figure \ref{fig:softmax}).

\begin{proposition}[Feed-forward neural network output]
    Let us given a feed-forward neural network $\mathcal{N}$ (see Definition \ref{def:1}) and $x \in \mathbb{R}^d$ with $d=card(\mathcal{I})$ then
    $$ \mathcal{S_N}(x) := \{x_v\}_{v \in \mathcal{O}} $$
    is a function mapping $\mathbb{R}^d$ into $\mathbb{R}^p$ where $p=card(\mathcal{O})$, called the output of $\mathcal{N}$.
\end{proposition}

The performance of these architectures is evaluated on new draws of data in the sense of probability or expectation of accuracy on said draws. Note that in this setting, there is no model class assumption on the function $f$ giving rise to the data, and so there can be no provable bound for the error. What is done in practice is to give an empirical bound based on checking performance from a lot of new random draws which are referred to as test data.

We can now consider fully-connected feed-forward neural networks.

\begin{proposition}[Fully-connected Neural networks]
\label{def:fully}
Let a feed-forward neural network $\mathcal{N}$ (see Definition \ref{def:1}) where the vertices are organized into layers. Each vertex of one layer is connected via outgoing edges to all vertices from the next layer and to no other vertices from any other layer.
If $X^{(l)} \in \mathbb{R}^{N^{(l)}}$ is the vector of neuron outputs corresponding to nodes $v$ in layer $l$ with $N^{(l)}=card(\{ v \in  \mathcal{V} \text{ ; v belonging to layer } l\})$, $l=0,...,L$ then the output of $\mathcal{N}$ is given by 
$$
\mathcal{S^{F}_N}(X^{(0)}) = \sigma^{(L)} \circ a^{(L)} \circ \sigma^{(L-1)} \circ a^{(L-1)}\circ...\circ \sigma^{(0)} \circ a^{(0)}(X^{(0)})
$$
where $X^{(0)}$ is made from all the vertices belonging to $\mathcal{I}$, $a^{(l)}(y)=W^{(l)} y +b^{(l)}$ being the linear activation vector of layer $l$ and $W^{(l)} \in \mathbb{R}^{N^{(l)} \times N^{(l-1)}}$ (resp. $b^{(l)} \in \mathbb{R}^{N^{(l)}}$) is made of each weight values (resp. bias values) issued from edges terminating in each vertex of layer $l$. The $\sigma^{(l)}$ activation function is defined to act on any vector coordinatewise.
\end{proposition}

When all the activation functions are the ReLU function, we then obtain Continuous Piecewise Linear (CPwL) functions \cite{DeVore2022} in the outputs.

\begin{figure}[t]
	\centering
    \begin{tikzpicture}[shorten >=1pt]
        \tikzstyle{unit}=[draw,shape=circle,minimum size=0.5cm, font=\tiny]

        \node[unit,fill=green!30](x000) at (-5.5,8.75){$x^{2l-1}_{00}$};
        \node[unit,fill=green!30](x100) at (-5.5,6.75){$x^{2l-1}_{10}$};
        \node[unit,fill=green!30](x200) at (-5.5,4.75){$x^{2l-1}_{20}$};
        \node[unit,fill=green!30](x300) at (-5.5,2.75){$x^{2l-1}_{30}$};
        \node[unit,fill=green!30](x400) at (-5.5,0.75){$x^{2l-1}_{40}$};

        \node[unit,fill=green!50](x010) at (-4,8.5){$x^{2l-1}_{01}$};
        \node[unit,fill=green!50](x110) at (-4,6.5){$x^{2l-1}_{11}$};
        \node[unit,fill=green!50](x210) at (-4,4.5){$x^{2l-1}_{21}$};
        \node[unit,fill=green!50](x310) at (-4,2.5){$x^{2l-1}_{31}$};
        \node[unit,fill=green!50](x410) at (-4,0.5){$x^{2l-1}_{41}$};

        \node[unit,fill=green!80](x020) at (-2.5,8.25){$x^{2l-1}_{02}$};
        \node[unit,fill=green!80](x120) at (-2.5,6.25){$x^{2l-1}_{12}$};
        \node[unit,fill=green!80](x220) at (-2.5,4.25){$x^{2l-1}_{22}$};
        \node[unit,fill=green!80](x320) at (-2.5,2.25){$x^{2l-1}_{32}$};
        \node[unit,fill=green!80](x420) at (-2.5,0.25){$x^{2l-1}_{42}$};

        \draw (-1.5,10) rectangle (-1,-1);
        
        \node[unit,fill=blue!30](x001) at (0,9){$x^{2l}_{00}$};
        \node[unit,fill=blue!30](x101) at (0,7){$x^{2l}_{10}$};
        \node[unit,fill=blue!30](x201) at (0,5){$x^{2l}_{20}$};
        \node[unit,fill=blue!30](x301) at (0,3){$x^{2l}_{30}$};
        \node[unit,fill=blue!30](x401) at (0,1){$x^{2l}_{40}$};

        \node[unit,fill=blue!60](x011) at (0.5,8){$x^{2l}_{01}$};
        \node[unit,fill=blue!60](x111) at (0.5,6){$x^{2l}_{11}$};
        \node[unit,fill=blue!60](x211) at (0.5,4){$x^{2l}_{21}$};
        \node[unit,fill=blue!60](x311) at (0.5,2){$x^{2l}_{31}$};
        \node[unit,fill=blue!60](x411) at (0.5,0){$x^{2l}_{41}$};

        \draw (2,9) rectangle (2.5,5.8);
        \draw (2,5.6) rectangle (2.5,2.8);
        \draw (2,2.6) rectangle (2.5,-0.2);

        \node[unit,fill=brown!30](x002) at (3,8){$x^{2l+1}_{00}$};
        \node[unit,fill=brown!30](x102) at (3,5){$x^{2l+1}_{10}$};
        \node[unit,fill=brown!30](x202) at (3,2){$x^{2l+1}_{20}$};
        
        \node[unit,fill=brown!60](x012) at (3.5,6.5){$x^{2l+1}_{01}$};
        \node[unit,fill=brown!60](x112) at (3.5,3.5){$x^{2l+1}_{11}$};
        \node[unit,fill=brown!60](x212) at (3.5,0.5){$x^{2l+1}_{21}$};

        \draw[->,color=blue!30] (x000) to[out=25, in=160] (x001) {};
        \draw[->,color=blue!60] (x000) to[out=25, in=160] (x011) {};
        \draw[->,color=blue!30] (x010) to[out=35, in=180] (x001) {};
        \draw[->,color=blue!60] (x010) to[out=35, in=180] (x011) {};
        \draw[->, color=blue!30] (x020) to[out=0, in=200] (x001) {};
        \draw[->,color=blue!60] (x020) to[out=0, in=200] (x011) {};
        
        \draw[->,color=blue!30] (x100) to[out=25, in=160] (x101) {};
        \draw[->,color=blue!60] (x100) to[out=25, in=160] (x111) {};
        \draw[->,color=blue!30] (x110) to[out=35, in=180] (x101) {};
        \draw[->,color=blue!60] (x110) to[out=35, in=180] (x111) {};
        \draw[->, color=blue!30] (x120) to[out=0, in=200] (x101) {};
        \draw[->,color=blue!60] (x120) to[out=0, in=200] (x111) {};

        \draw[->,color=blue!30] (x200) to[out=25, in=160] (x201) {};
        \draw[->,color=blue!60] (x200) to[out=25, in=160] (x211) {};
        \draw[->,color=blue!30] (x210) to[out=35, in=180] (x201) {};
        \draw[->,color=blue!60] (x210) to[out=35, in=180] (x211) {};
        \draw[->, color=blue!30] (x220) to[out=0, in=200] (x201) {};
        \draw[->,color=blue!60] (x220) to[out=0, in=200] (x211) {};

        \draw[->,color=blue!30] (x300) to[out=25, in=160] (x301) {};
        \draw[->,color=blue!60] (x300) to[out=25, in=160] (x311) {};
        \draw[->,color=blue!30] (x310) to[out=35, in=180] (x301) {};
        \draw[->,color=blue!60] (x310) to[out=35, in=180] (x311) {};
        \draw[->, color=blue!30] (x320) to[out=0, in=200] (x301) {};
        \draw[->,color=blue!60] (x320) to[out=0, in=200] (x311) {};

        \draw[->,color=blue!30] (x400) to[out=25, in=160] (x401) {};
        \draw[->,color=blue!60] (x400) to[out=25, in=160] (x411) {};
        \draw[->,color=blue!30] (x410) to[out=35, in=180] (x401) {};
        \draw[->,color=blue!60] (x410) to[out=35, in=180] (x411) {};
        \draw[->, color=blue!30] (x420) to[out=0, in=200] (x401) {};
        \draw[->,color=blue!60] (x420) to[out=0, in=200] (x411) {};

        \draw[->,color=blue!30] (x001) to[out=0, in=160] (x002) {};
        \draw[->,color=blue!30] (x101) to[out=0, in=180] (x002) {};
        \draw[->,color=blue!30] (x201) to[out=0, in=200] (x002) {};
        \draw[->,color=blue!60] (x011) to[out=0, in=160] (x012) {};
        \draw[->,color=blue!60] (x111) to[out=0, in=180] (x012) {};
        \draw[->,color=blue!60] (x211) to[out=0, in=200] (x012) {};

        \draw[->,color=blue!30] (x101) to[out=0, in=160] (x102) {};
        \draw[->,color=blue!30] (x201) to[out=0, in=180] (x102) {};
        \draw[->,color=blue!30] (x301) to[out=0, in=200] (x102) {};
        \draw[->,color=blue!60] (x111) to[out=0, in=160] (x112) {};
        \draw[->,color=blue!60] (x211) to[out=0, in=180] (x112) {};
        \draw[->,color=blue!60] (x311) to[out=0, in=200] (x112) {};

        \draw[->,color=blue!30] (x201) to[out=0, in=160] (x202) {};
        \draw[->,color=blue!30] (x301) to[out=0, in=180] (x202) {};
        \draw[->,color=blue!30] (x401) to[out=0, in=200] (x202) {};
        \draw[->,color=blue!60] (x211) to[out=0, in=160] (x212) {};
        \draw[->,color=blue!60] (x311) to[out=0, in=180] (x212) {};
        \draw[->,color=blue!60] (x411) to[out=0, in=200] (x212) {};

        \draw[decorate, decoration={brace, amplitude=1ex, raise=1cm}] (-2.5,0) -- node[midway, below=1.2cm] {$K_{2l-1}$} (-5.5,0);
        \draw[decorate, decoration={brace, amplitude=1ex, raise=1cm}] (x400.south) -- node[midway, left=1.2cm] {$I_{2l-1}=I_{2l}$} (x000.north);
        \draw[decorate, decoration={brace, amplitude=1ex, raise=1cm}] (1,0) -- node[midway, below=1.2cm] {$K_{2l}$} (-0.5,0);
        \draw[decorate, decoration={brace, amplitude=1ex, raise=1cm}] (4,0) -- node[midway, below=1.2cm] {$K_{2l+1}=K_{2l}$} (2.5,0);
        \draw[decorate, decoration={brace, amplitude=1ex, raise=1cm}] (3.5,8) -- node[midway, right=1.2cm] {$I_{2l+1}$} (3.5,0.5);
        
        \draw  (-1.5,10) -- node[midway, above=0.1cm] {$W^{(2l)}$} (-1,10);

        \draw  (2,9) -- node[midway, above=0.1cm] {$W^{2l+1}$} (2.5,9);
    \end{tikzpicture}
	\caption[Conv]{Convolution part implementation example in 1D}
	\label{fig:conv}
\end{figure}
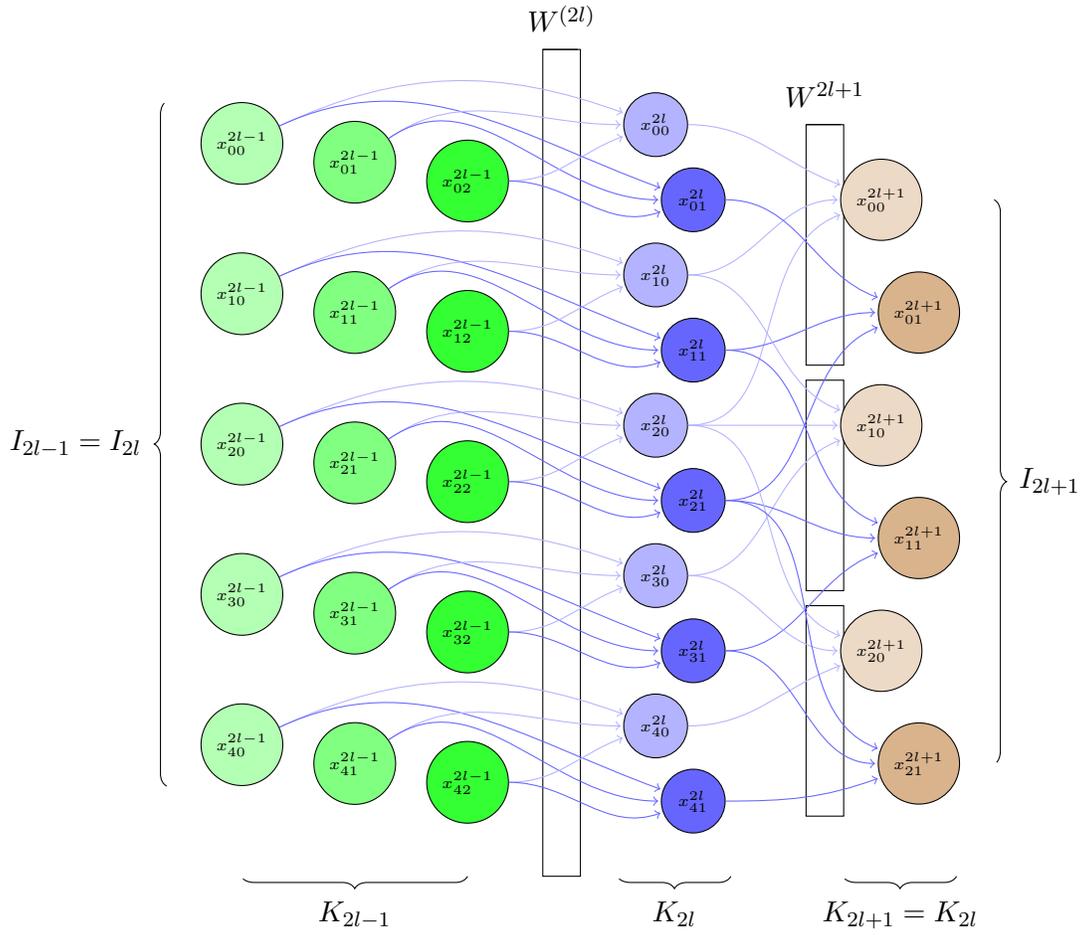

\begin{proposition}[Convolutional Neural networks part] 
\label{def:conv}
Let a feed-forward neural subnetwork $\mathcal{N}$ (see Definition \ref{def:1}) where the vertices are organized in $K_l$ grids of size $I_l$ into 2L layers. Therefore, each vertex $v_{(i,k)}$ has a coordinates (i, k) in a layer $l$, $i \in I_l$ being the position of the vertex in a grid, $k \in K_l$ being the grid number (figure \ref{fig:conv}).
Each vertex $i$ of one grid of one even layer $2l+2$ is connected via incoming edges to each vertex $i$ in all grid from the previous odd layer $2l+1$ and to no other vertices from any other previous layer. Each vertex $(i,k)$ of one grid of one odd layer 2l+1 is connected via incoming edges to each vertex $(i+m+P_l,k )$, $\forall m \in \mathbb{N}^{I_l} / m \leq 2M_l$ with $M_l \in \mathbb{N}^{I_l}$ and $P_l \in \mathbb{N}^{I_l}$, from the previous even layer $2l$ and to no other vertices from any other previous layer. 
If $X^{(l)} \in \mathbb{R}^{I_l \times K_l}$ is the tensor of neuron outputs corresponding to nodes $v_{(i,k)}$ in a layer $l$, $l=0,...,2L$ then the output of $\mathcal{N}$ is given by 
$$
\mathcal{S^{C}_N}(X^{(0)}) = \sigma^{(2L-1)} \circ a^{(2L-1)} \circ a^{(2L-2)}\circ...\circ \sigma^{(1)} \circ a^{(1)} \circ a^{(0)}(X^{(0)})
$$
where $X^{(0)} \in \mathbb{R}^{I_0 \times K}$ is made from all the vertices belonging to $\mathcal{I}$. $a^{(2l)}(y)=W^{(2l)} y +b^{(2l)}$ is the activation of layer $2l$ and $W^{(2l)} \in \mathbb{R}^{I_{2l} \times K_{2l} \times K_{2l-1} }$ (resp. $b^{(2l)} \in \mathbb{R}^{I_{2l} \times K_{2l} }$) is made of each weight values (resp. bias values) issued from edges terminating in each vertex of layer $2l$. $a^{(2l+1)}(y)=W^{(2l+1)} y +b^{(2l+1)}$ is the activation of layer $2l+1$ and $W^{(2l+1)} \in \mathbb{R}^{(1+2M_{2l+1}) \times K_{2l+1}} $ (resp. $b^{(2l+1)} \in \mathbb{R}^ {K_{2l+1}}$) which defines the weight values (resp. bias values) of edges terminating in each vertex of layer $2l+1$. The relationships between the layers are:$I_{(2l)}=I_{(2l-1)}$, $I_{(2l+1)}=I_{(2l)}-M_{(2l+1)}$ and $K_{(2l+1)}=K_{(2l)}$
\end{proposition}
In practice, a convolutional layer brings together one even layer followed by one odd layer.
Note that Average pooling layers and Max pooling layers are special cases of convolutional layers.
\begin{proposition}[Sequential Convolutional Neural networks]
\label{def:vgg}
A Sequential Convolutional Neural networks $\mathcal{N}$ such as VGG16 or VGG19, is defined by a Convolutional Neural networks part, followed by a Flatten layer if necessary, and after by a Fully-connected Neural networks. The Flatten layer allows to convert the tensor from the Convolutional part to a vector which can be used as input in the Fully-connected network. The output of $\mathcal{N}$ is then given by
$$
\mathcal{S^{CF}_N}(X^{(0)}) = \mathcal{S^F_N} \circ \mathcal{F}latten \circ \mathcal{S^{C}_N}(X^{(0)})
$$
where $X^{(0)} \in \mathbb{R}^{I_0 \times K}$ is made from all the vertices belonging to $\mathcal{I}$ and $\mathcal{F}latten$ is a Flatten layer.
\end{proposition}
\begin{proposition}[ResNet-Block type Neural networks] 
\label{def:resnet}
In \cite{He2016}, authors used residual learning blocks (see figure \ref{fig:skip}) in order to avoid optimization problem when training.  They defined a building block by $y=\mathcal{F}(x)+x$ where $x$ is the block input and $y$ the output. If $x$ and $y$ have the same size, a simple sum is done at the end of the block, otherwise a linear projection $W$ is performed onto $x$ before summing with $y$.

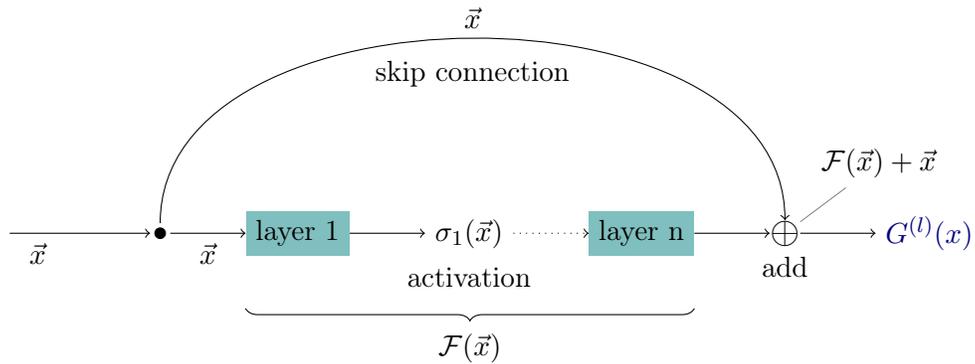
\begin{figure}[!h]  
  \centering
    \begin{tikzpicture}
        \node[] (act1) {$\pgfuseplotmark{*}$};
        \node[fill=teal!50, right=of act1] (l1) {layer 1};
        \node[label={below:activation}, right=of l1] (act2) {$\sigma_1(\vec x)$};
        \node[fill=teal!50, right=of act2] (ln) {layer n};

        \node[right=of ln, font=\Large, label={below:add}, inner sep=0, pin={60:$\mathcal F(\vec x) + \vec x$}] (add) {$\oplus$};

        \node[blue!50!black, right=of add] (out) {$G^{(l)}(x)$} ;

        \draw[<-] (act1) -- ++(-2,0) node[below, pos=0.8] {$\vec x$};
        \draw[->] (act1) -- (l1) node[below, pos=0.5]{$\vec x$};
        \draw[->] (l1) -- (act2);
        \draw[->,dotted] (act2) -- (ln);
        \draw[->] (ln) -- (add) node[above, pos=0.8] {};
        \draw[->] (add) -- (out) node[below, pos=0.8] {};
        \draw[->] (act1) to[out=90, in=90] node[below=1ex, midway, align=center] {skip connection} node[above, midway] {$\vec x$} (add);
        \draw[decorate, decoration={brace, amplitude=1ex, raise=1cm}] (ln.east) -- node[midway, below=1.2cm] {$\mathcal F(\vec x)$} (l1.west);
    \end{tikzpicture}
    \caption{A residual learning building block.}
    \label{fig:skip}
\end{figure}

As $\mathcal{F}$ is a Convolutional Neural Networks Part, for any input vector ${\bf X}^{(0)}$, the output of a ResNet Neural network can be defined by 
    $$
        \mathcal{S^R_N}(X^{(0)}) =  \mathcal{S^F_N} \circ \mathcal{F}latten \circ \mathcal{S^{C}_N}_{(q-1)}  \circ \sigma^{(q-2)} \circ G^{(q-2)}\circ... \sigma^1 \circ G^{(1)}\circ \mathcal{S^{C}_N}_{(0)}(X^{(0)})
    $$
    where 
    $G^{(l)}(x)=\mathcal{S^{C}_N}_{(l)}(x) +W_{(l)}x$ is a ResNet-Block
\end{proposition}

\subsection{LRP background}
The LRP methods decompose the output of a nonlinear decision function in terms of input variables, forming a vector of input features scores (i.e. contributions) that constitutes the 'explanation’. For the decomposition process, LRP methods usually use back propagation (from one network output to its inputs) in order to calculate a contribution score of each input.
There are different back propagation rules for LRP. For LRP0 method, the contribution is defined in the following form:
\begin{definition}[LRP0]
\label{def:LRP0}
According to the definition in \cite{Bach2015}, the contribution of neuron $i$ of layer $l$ to the activation/output $x_j^{(l+1)}$ of neuron $j$ from layer $l+1$ is given by
\begin{equation}
\label{eq:1}
    {\bf z}^{(l)}_{i,j}=\frac {w_{i,j}^{(l)} {\bf x}_i^{(l)}}
            { {\displaystyle \sum_{k}{w_{k,j}^{(l)}} {\bf x}_k^{(l)}}} {\bf x}_j^{(l+1)}
\end{equation}
\end{definition}

 Resulting scores can be visualized as a heatmap of the same dimensions as the input. Initially, LRP is a conservative technique, meaning the magnitude of any output is conserved through the backpropagation process and is equal to the sum of the relevance map of the input layer ; this is call conservation law \cite{Bach2015}. The property holds for any consecutive layers, and by transitivity for the input and output layer.

\begin{definition}[Conservation Law] 
\label{def:law}
According to the definition in \cite{Bach2015}, the contribution law states that 
\begin{equation}
\label{eq:law}
    \sum_{N^{(l)}}{\it z}_{p}^{(l)} = x^{(q+1)} \text{\ for\ all\  } l \in \{0,q\}
\end{equation}
where  $q$ is the depth of the neural network, ${\it z}^{(l)}=({\bf z}^{(l)}_{i,j})_{i,j\in l,l+1}$ of length $N^{(l)}$ is a vectorization of ${\bf z}^{(l)}$ and $x^{(q+1)}$ is one selected output value. 
\end{definition}

\subsection{LRP0 drawback}
\begin{figure}[!h]  
  \centering
       \begin{tabular}{|c|>{\centering\arraybackslash}m{1.5cm}|>{\raggedright\arraybackslash}m{1.5cm}|>{\centering\arraybackslash}m{1.5cm}|>{\centering\arraybackslash}m{1.5cm}|>{\centering\arraybackslash}m{1.5cm}|>{\centering\arraybackslash}m{1.5cm}|}
        \hline
        Image& & 1\% & 5\% & 10\% & 20\% & 50\%\\
        \hline
        \multirow{ 2}{*} {\includegraphics[width=1.45cm]{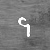} } &
         Mask &
        \includegraphics[width=1.45cm]{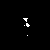} & 
        \includegraphics[width=1.45cm]{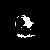}& 
        \includegraphics[width=1.45cm]{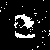} & \includegraphics[width=1.45cm]{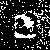}& \includegraphics[width=1.45cm]{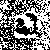} \\ \cline{2-7}
        & Masked Image &
        \includegraphics[width=1.5cm]{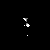} & 
        \includegraphics[width=1.5cm]{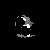}& 
        \includegraphics[width=1.5cm]{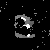} & \includegraphics[width=1.5cm]{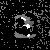}& \includegraphics[width=1.5cm]{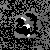}\\
        \hline
        \end{tabular}
        
  \caption{Images showing percentage of the most relevant pixels using LRP0 on a modified MNIST dataset using a simple dense network. Mask images are in black and white}
    \label{fig:1}
\end{figure}

The problem of the LRP0 method is mainly due to the fact that the calculation of the contribution $z_{i,j}$ (see above) is numerically very sensitive to the value of $ \sum_{k}{w_{k,j}^{(l)} {\bf x}_k^{(l)}}$, as we can see in figure \ref{fig:1}, for some pixels of layer $l$ which do not contribute a lot to the final classification. The LRP0 formula (1) may admit small denominators $\sum_{k}{w_{k,j}^{(l)}} {\bf x}_k^{(l)} \simeq 0$. Consequently, the surrounding pixel contributions in this layer $l$ can be amplified positively and negatively. Hence, their contributions become non-representative \cite{Bach2015}. 

To illustrate this effect, we have trained a simple network composed of three dense layers with respectively 256, 128 and 10 neurons using a modified MNIST dataset with 50x50 images where  the background is made of non-uniform gray values. After 50 epochs, the network reached an accuracy of 0.925 for the training dataset and 0.91 for the validation dataset. The LRP0 is then applied to it to evaluate the effectiveness of the method. As we can see in figure \ref{fig:1}, the contributions obtained are not so relevant.

\begin{figure}[!h]  
  \centering
       \begin{tabular}{|>{\centering\arraybackslash}m{1.5cm}|>{\centering\arraybackslash}m{1.5cm}|>{\centering\arraybackslash}m{1.5cm}|>{\centering\arraybackslash}m{1.5cm}|>{\centering\arraybackslash}m{1.5cm}|>{\centering\arraybackslash}m{1.5cm}|}
        \hline
        Image& 1\% & 5\% & 10\% & 20\% & 50\%\\
        \hline
        \includegraphics[width=1.5cm]{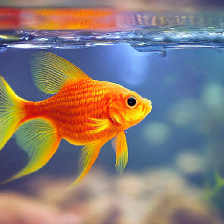} & 
        \includegraphics[width=1.5cm]{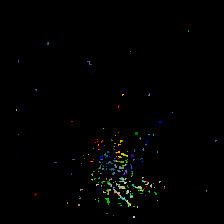}& 
        \includegraphics[width=1.5cm]{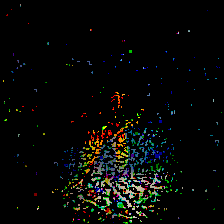}& 
        \includegraphics[width=1.5cm]{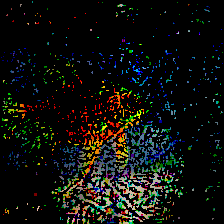} & \includegraphics[width=1.5cm]{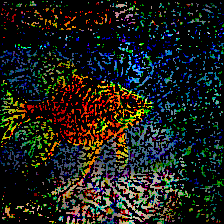}& \includegraphics[width=1.5cm]{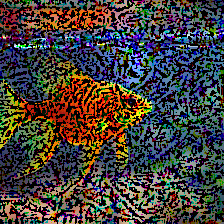}\\
        \hline
        \includegraphics[width=1.5cm]{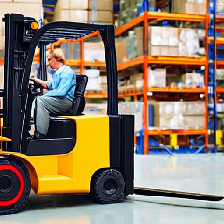}  &
        \includegraphics[width=1.5cm]{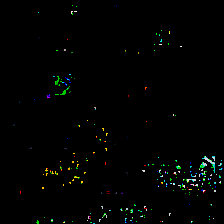} & 
        \includegraphics[width=1.5cm]{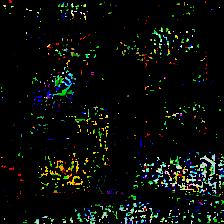}& 
        \includegraphics[width=1.5cm]{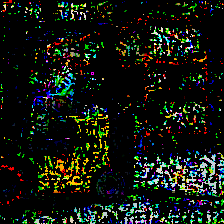} & \includegraphics[width=1.5cm]{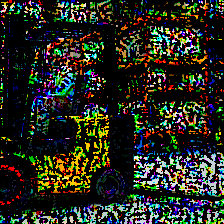}& \includegraphics[width=1.5cm]{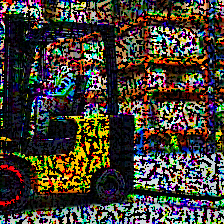} \\ 
        
        \hline
        \end{tabular}
        
  \caption{Images showing percentage of the most relevant inputs using LRP0 on two images from Imagenet dataset with VGG16.}
    \label{fig:2}
\end{figure}

With deeper CNN networks, for example a Keras-VGG16 network whose coefficients are pre-trained on ImageNet and RGB images,
the results without any post-processing step is quite disappointing (see figure \ref{fig:2}). Consequently, direct LRP0 calculations for a given network do not provide a representative image of contributions. 

\subsection{LRP modified formulas}
In order to avoid this problem, some LRP-methods use modified formulas to finally produce post processed heatmaps. Usually, these methods are combined to get the best results. For images, the rule of choice has been introduced in \cite{Montavon2017}. For lower-layers, the explanation has to be more smooth and less noisy since these layers are already very close to the relevance map we humans will see and have to make sense of. For this purpose, LRP-$\gamma$ (see equation (\ref{eq:2})) rule that disproportionately favors positive evidence over negative evidence is used:

\begin{equation}
\label{eq:2}
z^{(l)}_{i}=\sum_{j} {\frac {(w_{i,j}^{(l)}+\gamma w_{i,j}^{(l)+}){\bf x}_i^{(l)}}
            { {\displaystyle \sum_{k} (w_{k,j}^{(l)}+\gamma w_{k,j}^{(l)+}) {\bf x}_k^{(l)}}}}{\bf x}_j^{(l+1) }
\end{equation}
where $x^+$ denotes the function $max(0,x)$.

For higher-layers, either LRP-$\gamma$ or LRP-$\epsilon$ rule are used. They remove some of the noise from the relevance map. In particular, LRP-$\epsilon$ (see equation (\ref{eq:3})) rule aims to solve the problem of gradient noise by introducing a small positive term, $\epsilon$, to the denominator in order to absorb some relevance when contributions are weak or contradictory.
\begin{equation}
\label{eq:3}
z^{(l)}_{i}=\sum_{j} {\frac {w_{i,j}^{(l)}{\bf x}_i^{(l)}}
            { {\displaystyle \epsilon +\sum_{k}w_{k,j}^{(l)} {\bf x}_k^{(l)}}} {\bf x}_j^{(l+1)}}
\end{equation}

In \cite{Bach2015}, LRP-$\alpha\beta$ rule is used to treat positive and negative contributions asymmetrically. This alternate propagation method also allows controlling
manually the importance of positive and negative evidence, by choosing different factors $\alpha$ and $\beta$ with $\alpha+\beta=1$: 
\begin{equation}
\label{eq:4}
z^{(l)}_{i}=\alpha \sum_{j} {\frac {(w_{i,j}{\bf x}_i^{(l)})^{+}}
            { {\displaystyle \sum_{k} (w_{k,j}{\bf x}_k^{(l)})^{+}}} {\bf x}_j^{(l+1) }} +
            \beta \sum_{j} {\frac {(w_{i,j}{\bf x}_i^{(l)})^{-}}
            { {\displaystyle \sum_{k} (w_{k,j}{\bf x}_k^{(l)})^{-}}} {\bf x}_j^{(l+1) }}
\end{equation}
where $x^{-}$ denotes $min(0,x)$.

In \cite{Hu2021}, the authors introduced LRP-$z^{+}$ or LRP-$w^{2}$ which give similar results than LRP-$\alpha\beta$ and LRP-$\gamma$.

Conservation law (see Definition \ref{def:law}) for the global relevance in between layers can only be guaranteed with LRP-$\gamma$ and LRP-$\alpha\beta$ when $\alpha+\beta=1$.
In \cite{Bach2015}, authors used $\beta=-1$ and so $\alpha=2$ in order to get usable heatmaps as they assume that "for rectified linear neurons this is a reasonable assumption because they fire only when their input is positive". 
The main issue of these modified methods is that we have to adjust hyperparameters ($\gamma, \epsilon, \alpha, \beta$, ...) and/or to mix these methods depending on the dataset.

\section{Relative LRP or R-LRP method}
\label{LRPR}

\subsection{Introduction}
In this section, we introduce new relative input/pixel contributions. We call this method R-LRP where "R" stands for Relative.
The objective is to have no hyperparameter to tune and no need to mix LRP methods.

As mention above, equation (\ref{eq:1}) doesn't give a representative relative magnitude of the contributions when $\sum_{k}{w_{k,j}^{(l)}} {\bf x}_k^{(l)} \simeq 0$. Indeed, in these cases, input contributions  $z_{i,j}^{(l)}$ may have huge values whereas outputs ${\bf x}_j^{(l+1)}$ are close to zero. To avoid this phenomenon, we propose this new backward contribution rule definition. 
\begin{definition}[R-LRP]
     According to Definition \ref{def:1}, let us give the output $k$ of the final layer then the contribution of neuron $i$ of layer $l$ to the output $x_j^{(l+1)}$ of neuron $j$ from layer $l+1$ is defined by
    \begin{equation}
    \label{eq:5}
      z_{i,j,k}^{(l)} = \frac {1}{M_j^{(l+1)}}w_{i,j}^{(l+1)}  {\bf x}_i^{(l)}{\bf z}_{j,k}^{(l+1)} 
    \end{equation}
    where $M_j^{(l+1)}$ is the input number of neurons connected to neuron $j$.
    We then define the relative contribution of neuron $i$ from layer $l$ on any neuron output $k$ of the final layer by
    \begin{equation}
    \label{eq:6}
      {\bf z}_{i,k}^{(l)} =  \frac {Card(J)}{N^{(l+1)}}\sum_{J}z_{i,j,k}^{(l)} 
    \end{equation}
    where $N^{(l+1)}$ is the number of neurons in layer $l+1$ and $J$ being the set of all neuron indexes from layer $l+1$ connected to neuron $i$ from layer $l$.
    We set ${\bf z}_{k,k}^{(q+1)}:={\bf x}_k^{(q+1)}$ for any neuron $k$ in the final layer. Consequently, $z_{i,k}^{(0)}$ is the contribution of the input/pixel $i$ on the output neuron $k$ of the final layer.
\end{definition}

The main advantage of this formula is that there is no division by small values, nor any hyperparameter to set or tune. The division by the number of neurons in the current layer avoids some overflow problems but does not affect relative contributions.

The total contribution of an input to an output, up to a factor that is the same for all inputs, is calculated independently of other inputs. For networks which can be associated with a DAG, the sum of the inputs' contributions is equal to the selected output (up to a factor for our case). Therefore, up to a real factor we respect the classical LRP-methods conservation law (see Equation \ref{eq:law}) because an input contribution is relative to an output value and thus relative to other inputs' contributions. In this sense, the contributions are relative. Moreover, calculations are lighter and, therefore, faster.

\begin{proposition}[R-LRP contribution and activation]
    According to the previous definition, R-LRP only keeps input which do not inactivate neuron(s) in the network. The activation threshold of a neuron $l_j^{(l+1)}$ is specified by the bias term, thus, equation (\ref{eq:5}) makes it possible to take into account the contribution of neuron although $\sum_{k}{w_{k,j}^{(l)}} {\bf x}_k^{(l)} = 0$
    
\end{proposition}

\subsection{Effective R-LRP calculations from CNN layer components}
In practice, efficient calculations of the contributions takes into account the layer component in order to simplify/rearrange equation \ref{eq:6}

\begin{proposition}[R-LRP Dense]
In the case of Dense layers, the contribution of a neuron $i$ for the output $k$ can be calculated by
    \begin{equation}
    \label{eq:7}
      {\bf z}_{i,k}^{(l)} =  \frac {1}{N^{(l)}}\sum_{j\in J}w_{i,j}^{(l+1)}{\bf x}_i^{(l)}{\bf z}_{j,k}^{(l+1)}
    \end{equation}
    where $Card(J)=N^{(l+1)}$ and $N^{(l)}=M_j^{(l+1)}$.
\end{proposition}

\begin{proposition}[R-LRP Convolution]
A convolution layer with pool-size ($P_1,P_2$), filter mask $W=\{w_{(p_1,p_2)}\setminus p_1 \in P_1 , p_2 \in P_2\}$ and strides ($s_1,s_2$) acts like a grid of neurons where each neuron $(j_1,j_2)$ have $P_1xP_2$ entries (see Definition \ref{def:conv}) and the output are
\begin{equation}
    {\bf x}_{(j_1,j_2)}^{(l+1)} = \sum_{p_1}\sum_{p_2} w_{(p_1,p_2)}.{\bf x}_{(j_1.s_1+p_1),(j_2.s_2+p_2)}^{(l)}
\end{equation}
where the values ${\bf x}_{(j_1.s_1+p_1),(j_2.s_2+p_2)}^{(l)}$ are neuron/pixel contribution from layer $l$.

Thus, from equation (\ref{eq:5}), the contribution of a neuron/pixel ${(i_1,i_2)}$ of layer $l$ to the output ${\bf x}_{(j_1,j_2)}^{(l+1)}$ of the neuron ${(j_1,j_2)}$ for the final output $k$ is defined by
\begin{equation}
{\bf z}_{(i_1,i_2),(j_1,j_2),k}^{(l)} =\frac{1}{P_1.P_2} w_{(p_1,p_2)} {\bf x}_{(i_1,i_2)}^{(l)} {\bf z}_{(j_1,j_2),k}^{(l+1)}
\end{equation}
where $i_1=j_1.s_1+p_1$ and $i_2=j_2.s_2+p_2$
The total contribution of a neuron/pixel ${(i_1,i_2)}$ of layer $l$ for the final output $k$ is obtained from equation (\ref{eq:6}):
\begin{equation}
    \label{eq:10}
    {\bf z}_{(i_1,i_2),k}^{(l)} =  \frac {Card(J)}{N^{(l+1)}}\sum_{J}\frac{1}{P_1.P_2} w_{(p_1,p_2)} {\bf x}_{(i_1,i_2)}^{(l)} {\bf z}_{(j_1,j_2),k}^{(l+1)}
\end{equation}
Since, $\frac{1}{P_1.P_2}$ is a constant and ${\bf x}_{(i_1,i_2)}^{(l)}$ is independent of $J$, we can rewrite equation (\ref{eq:10}) as 
\begin{equation}
    \label{eq:11}
    {\bf z}_{(i_1,i_2),k}^{(l)} =\frac {Card(J)}{N^{(l+1)}}\frac{1}{P_1.P_2} {\bf x}_{(i_1,i_2)}^{(l)} \sum_{J} w_{(p_1,p_2)} {\bf z}_{(j_1,j_2),k}^{(l+1)}
\end{equation}
where $ \sum_{J} w_{(p_1,p_2)} {\bf z}_{(j_1,j_2),k}^{(l+1)}$ is calculated with a transposed convolution \cite{Dumoulin2016}.
This definition is also available for convolutions with padding.
\end{proposition}

\begin{proposition}[R-LRP Average and Max Pooling]
An Average Pooling2D layer with pool-size ($P_1,P_2$) is a convolution layer where the filter mask is defined by
$$W=\{w_{(p_1,p_2)}=\frac{1}{P_1.P_2} \ with \ p_1 \in P_1 , p_2 \in P_2\}.$$
Thus in case of pooling layers, we can use equation (\ref{eq:11}) with the corresponding filter mask.

A Max Pooling2D layer can also be considered as a convolution layer with a filter mask
$$W=\{w_{(p_1,p_2)}= {\bf 1}_{(k_1,k_2)} \ if\  (k_1,k_2)\in \argmaxD_{p_1 \in P_1 , p_2 \in P_2} {\bf x}_{(i_1.s1+p_1,i_2.s2+p_2)} \ \ otherwise \  0 \}.$$
As the filter mask depends on $(i_1,i_2)$, the calculation of \[ \sum_{J} w_{(p_1,p_2)} {\bf z}_{(j_1,j_2),k}^{(l+1)} \] in equation (\ref{eq:11}) cannot be obtained by a simple transposed convolution but by an adapted one. The argmax function returns all the indices of the maximum values in a list or in an array.
\end{proposition}
\subsection{R-LRP calculations with ResNet}
In order to get the contributions at the input of a block, one must calculate the contributions of the residual part ($\mathcal{S^{C}_N}(x)$), the contributions of the skip connection part ($W_sx$) and then sum the two contributions.
About ResNet, in \cite{He2016}, authors wrote that:
\begin{enumerate}
    \item if the input and output have the same size, a simple sum is done at the end of the block. So for the skip connexion part, the contributions of the input are the same as the output.
    \item if the output and input sizes are different, they perform a linear projection $W_s$ by the shortcut connection to match the dimensions : $y=\mathcal{S^{C}_N}(x)+W_sx$. Practically, the projection is done by a 1x1 convolution with a downsampling stride of 2. For the skip connexion part, the contributions of the input of the block are obtained by oversampling the output contributions using eq. \ref{eq:11} 
\end{enumerate}

For the residual part, as $\mathcal{S^{C}_N}(x)$ is a Convolutional part, R-LRP calculation is used. 

\begin{figure}[!h]  
  \centering
  \begin{tabular}{m{3cm} m{3cm}  m{3cm}}
  \includegraphics[width=3cm]{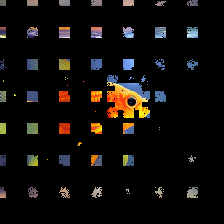}
  & &
  \includegraphics[width=3cm]{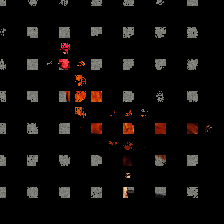}
  \end{tabular}
    \caption{R-LRP problems with non-normalized sums for residual block contributions}
    \label{fig:skip_contributions_nn}
\end{figure}

While summing the contributions, we must take care of the conservation law (definition \ref{def:law}) because if residual's and skip connection's contributions don't have the same magnitude, artifacts occur in the input contributions result (see Figure \ref{fig:skip_contributions_nn}). We can also see this problem in \cite[figure7]{Jee2021} where the artifacts are mostly due to the oversampling in the skip part.

R-LRP contributions are relatives, thus we have to "normalize" the residual and the skip connection contributions before summing them. From the conservation law (equation \ref{eq:law}), we can state  that, 
$$\sum_{N^{(l)}}{\it z}_{p}^{(l)}=\sum_{N^{(m)}}{\it z}_{p}^{(m)}, \forall (l,m)\in \{0,...,q\}\times\{0,...,q\}$$ 
Thus, the normalization consists in having the same value for sums at the end and the beginning of each part of the residual building block. This normalization ensures that no artifact occurs due to separate evaluation of each part of the residual block.

\begin{figure}[!h]  
  \centering
    \begin{tikzpicture}
        \node[] (act1) {$\pgfuseplotmark{*}$};
        \node[fill=teal!50, right=of act1] (l1) {layer 1};
        \node[ right=of l1] (act2) {$\sigma_1(\vec x)$};
        \node[fill=teal!50, right=of act2] (ln) {layer n};

        \node[right=of ln, font=\Large, inner sep=0] (add) {$\oplus$};

        \node[blue!50!black, right=of add] (out) {} ;

        \draw[<-] (act1) -- ++(-2,0) node[below, pos=0.8] {${\bf z}^{(l)}_{i,k}$};
        \draw[->] (act1) -- (l1) node[below, pos=0.5]{${\bf z_2}_{i,k}^{(l)}$};
        \draw[->] (l1) -- (act2);
        \draw[->,dotted] (act2) -- (ln);
        \draw[->] (ln) -- (add) node[below, pos=0.5] {${\bf z_2}_{i,k}^{(m)}$};
        \draw[->] (add) -- (out) node[below, pos=0.8] {${\bf z}^{(m)}_{i,k}$};
        \draw[->] (act1) to[out=90, in=90] node[near start,sloped,above] {${\bf z_1}_{i,k}^{(l)}$ } node[fill=teal!50,midway] {$W_sx$} node[near end,sloped,above] {${\bf z_1}_{i,k}^{(m)}$}(add);
        
    \end{tikzpicture}
    \caption{Contribution calculus for residual learning building block.}
    \label{fig:skip_contrib}
\end{figure}
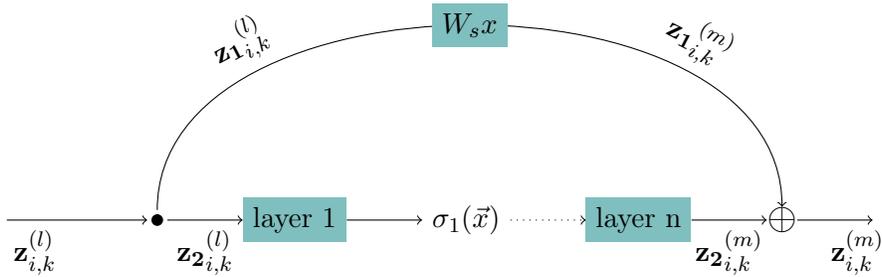

\begin{proposition}[R-LRP Residual Learning Block]
A learning block is defined by $$y=\mathcal{S^{C}_N}(x)+W_sx$$ where $\mathcal{S^{C}_N}(x)$ defines a residual part that uses CNN layer components and $W_sx$ is a skip connection with a convolution $W_s$. Based on the conservation law property, we then define the contribution of the inputs of a learning block by
\begin{equation}
\label{eq:8}
  {\bf z}_{i,j}^{(l)} =  {\bf z_1}_{i,k}^{(l)}\frac{\sum_{N^{(m)}}{\it z_1}_{p}^{(m)}} {\sum_{N^{(l)}}{\it z_1}_{p}^{(l)}}+ {\bf z_2}_{i,k}^{(l)}\frac{\sum_{N^{(m)}}{\it z_2}_{p}^{(m)}} {\sum_{N^{(l)}}{\it z_2}_{p}^{(l)}}
\end{equation}
where ${\bf z_1}_{i,k}^{(l)}$ (resp. ${\bf z_1}_{i,k}^{(m)}$) is the contribution at the beginning (resp. at the end) of the block using the skip connection part path ($W_sx$) and ${\bf z_2}_{i,k}^{(l)}$ (resp. ${\bf z_2}_{i,k}^{(m)}$) is the contribution at the beginning (resp. at the end) of the block using the residual part path ($\mathcal{S^{C}_N}(x)$) (see Figure \ref{fig:skip_contrib}). 
\end{proposition}

\section{Evaluations with MNIST-like dataset}
\label{MNIST}

\begin{figure}[!h]  
  \centering
       \begin{tabular}{|>{\centering\arraybackslash}m{1.5cm}|>{\raggedright\arraybackslash}m{1.3cm}|>{\raggedright\arraybackslash}m{1.8cm}|>
       {\centering\arraybackslash}m{1.2cm}|>{\centering\arraybackslash}m{1.2cm}|>{\centering\arraybackslash}m{1.2cm}|>{\centering\arraybackslash}m{1.2cm}|>{\centering\arraybackslash}m{1.2cm}|}
        \hline
        Image& method & parameters & 1\% & 5\% & 10\% & 20\% & 50\%\\
        \hline
        \multirow{ 4}{*} {\includegraphics[width=1.4cm]{images/data/mnist/LRP0_mnist_dense/image11cat_9_9_ori.png} } &
         LRP-$0$ & &
        \includegraphics[width=1.2cm]{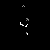} & 
        \includegraphics[width=1.2cm]{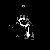}& 
        \includegraphics[width=1.2cm]{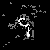} & \includegraphics[width=1.2cm]{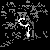}& \includegraphics[width=1.2cm]{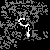} \\ \cline{2-7}
        & LRP-$\epsilon$ & $\epsilon=0.01$ &
        \includegraphics[width=1.2cm]{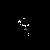} & 
        \includegraphics[width=1.2cm]{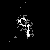}& 
        \includegraphics[width=1.2cm]{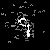} & \includegraphics[width=1.2cm]{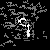}& \includegraphics[width=1.2cm]{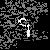}\\
        \cline{2-7}
        & LRP-$\gamma$ & $\gamma=0.25$ &
        \includegraphics[width=1.2cm]{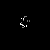} & 
        \includegraphics[width=1.2cm]{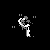}& 
        \includegraphics[width=1.2cm]{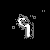} & \includegraphics[width=1.2cm]{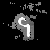}& \includegraphics[width=1.2cm]{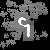}\\
        \cline{2-7}
        & LRP-$\alpha\beta$ & $\alpha=2$ $\beta=-1$ &
        \includegraphics[width=1.2cm]{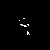} & 
        \includegraphics[width=1.2cm]{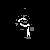}& 
        \includegraphics[width=1.2cm]{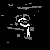} & \includegraphics[width=1.2cm]{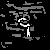}& \includegraphics[width=1.2cm]{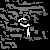}\\
        \cline{2-7}
        & R-LRP &  &
        \includegraphics[width=1.2cm]{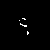} & 
        \includegraphics[width=1.2cm]{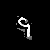}& 
        \includegraphics[width=1.2cm]{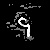} & \includegraphics[width=1.2cm]{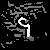}& \includegraphics[width=1.2cm]{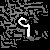}\\
        \hline
        \end{tabular}
        
  \caption{Images showing percentage of the most relevant pixels using LRP methods on a modified MNIST dataset using a simple CNN network.Positive contributions are in white while negative ones are in grey}
    \label{fig:3}
\end{figure}

In order to evaluate the LRP methods, we have designed a network (using Python and the Keras library) composed of two convolution layers ($32$ filters each), a flatten one and a $128$-dense layer before the output dense layer. After $5$ epochs training iterations with batch size of $128$, we obtained an accuracy of $98\%$.
For each image in the modified MNIST test dataset, we have created images corresponding to, respectively, 1\%, 5\%, 10\%, 15\%, 20\%, 25\%, 40\%, 50\%, 60\%, 75\%, 80\%, 85\%, 90\%, 95\%, 99\% of the most relevant pixels for each LRP method (see figure \ref{fig:3} for example). In this article, we only show results for test image dataset which are identified in their effective category because for train dataset images, we obtain similar results and for misidentified images the results are symmetrical.

\begin{figure}[!h]  
  \centering
  \includegraphics[width=10cm]{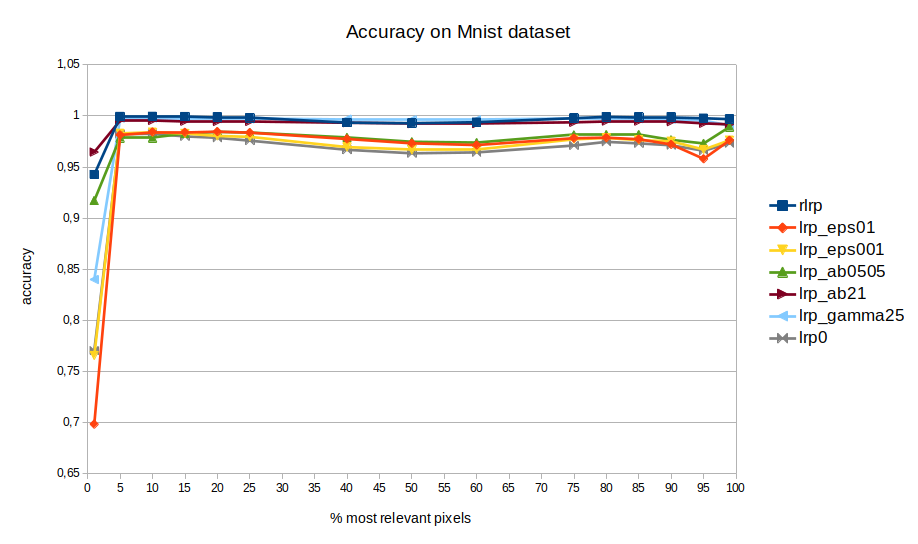}
    \caption{Accuracies of LRP methods over modified  MNIST test dataset in terms of percentage of the most relevant pixels }
    \label{fig:4}
\end{figure}

The evaluation gives the accuracy of the test set based on the percentage of the most relevant pixels in the images for each LRP methods with commonly used parameter values (see figure \ref{fig:4}). In all graphs, rlrp stands for R-LRP, lrp0 stands for LRP-0 method, lrp\_eps01 (resp. lrp\_eps001) stands for LRP-$\epsilon$ method with parameter $\epsilon$=0.01 (resp. $\epsilon$=0.001), lrp\_ab0505 (resp. lrp\_ab21) stands for LRP-$\alpha\beta$ method with parameters $\alpha$=0.5 and $\beta$=0.5 (resp. $\alpha$=2 and $\beta$=-1) and lrp\_gamma25 stands for LRP-$\gamma$ with parameter $\gamma$=0.25. Accuracy is calculated using the trained simple CNN network according to the founded category and the pixels of the images, whose contributions are less than the percentage, in black. We prefer this method because masking relevant pixels in images can induce a bias due to the shape of the mask itself and guide the choice of the network.

Although all these methods achieve quickly more than 95\% of success, we naturally want the contributions of pixels to focus on the object in the input image and therefore, will have a higher value on the corresponding input unlike the other pixels which must have lower contributions. With modified LRP formulas, this seems to not be completely the case (see \ref{fig:3}). Moreover, these methods need to adjust some hyperparameters ($\gamma, \epsilon, \alpha, \beta$, ...) which are image sets dependent and require human experts in order to be understandable \cite{Montavon2019}. 

\subsection{Using positive and negative contributions}
In figures \ref{fig:4}, we can see that in between about $20\%$  and $80\%$ of most relevant pixels, all the accuracies decrease. This seems to show that pixels with negative contributions are also significant. 

\begin{figure}[!h]  
  \centering
       \begin{tabular}{|>{\centering\arraybackslash}m{1.5cm}|>{\raggedright\arraybackslash}m{1.3cm}|>{\raggedright\arraybackslash}m{1.8cm}|>
       {\centering\arraybackslash}m{1.2cm}|>{\centering\arraybackslash}m{1.2cm}|>{\centering\arraybackslash}m{1.2cm}|>{\centering\arraybackslash}m{1.2cm}|>{\centering\arraybackslash}m{1.2cm}|}
        \hline
        Image& method & parameters & 1\% & 5\% & 10\% & 20\% & 50\%\\
        \hline
        \multirow{ 4}{*} {\includegraphics[width=1.4cm]{images/data/mnist/LRP0_mnist_dense/image11cat_9_9_ori.png} } &
         LRP-$\gamma$ & $\gamma=0.25$ &
        \includegraphics[width=1.2cm]{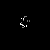} & 
        \includegraphics[width=1.2cm]{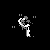}& 
        \includegraphics[width=1.2cm]{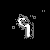} & \includegraphics[width=1.2cm]{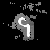}& \includegraphics[width=1.2cm]{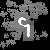} \\ \cline{2-7}
        & LRP-$\alpha\beta$ & $\alpha=2$ $\beta=-1$ & 
        \includegraphics[width=1.2cm]{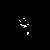} & 
        \includegraphics[width=1.2cm]{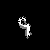}& 
        \includegraphics[width=1.2cm]{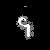} & \includegraphics[width=1.2cm]{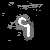}& \includegraphics[width=1.2cm]{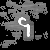}\\
        \cline{2-7}
        
        & R-LRP &  &
        \includegraphics[width=1.2cm]{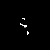} & 
        \includegraphics[width=1.2cm]{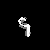}& 
        \includegraphics[width=1.2cm]{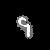} & \includegraphics[width=1.2cm]{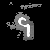}& \includegraphics[width=1.2cm]{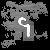}\\
        \hline
        \end{tabular}
        
  \caption{Images showing percentage of the most relevant pixels in absolute value using some LRP methods on a modified MNIST dataset using a simple CNN network.}
    \label{fig:7}
\end{figure}

Thus, we have evaluated our R-LRP methods using absolute value of the contributions. This means that the most relevant pixels in an image (see figure \ref{fig:7}) are defined by the absolute value of the contributions of each pixel calculated with equation (\ref{eq:5}).

\begin{figure}[!h]  
  \centering
  \includegraphics[width=10cm]{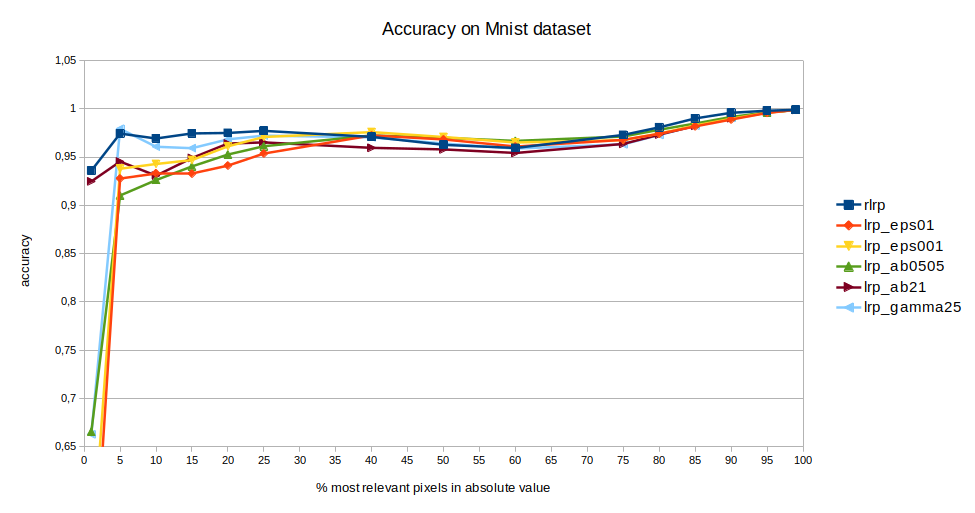}
    \caption{Accuracy of LRP methods over modified  MNIST test dataset in terms of percentage of the most relevant pixels in absolute value}
    \label{fig:8}
\end{figure}

In figure \ref{fig:8}, we can see that with 5\% of the most relevant pixels in absolute values, the accuracy reaches over than 98\% for each category.
Although the scores are worse with small percentage of most relevant pixels by using absolute value contributions than using direct contributions, it seems that pixels with negatives contributions are closest to the digit in the image (figure \ref{fig:7}).

This definitely shows the relevance of our R-LRP method to define the contribution of the pixels in the decision process with our simple "MNIST-network" (which contains dense and convolution layers) for MNIST dataset.

\section{R-LRP validation with cat vs dog dataset}
In order to validate our new formula \ref{eq:5}, we decided to use a more complex CNN than the one used for MNIST dataset and a more complex dataset. To do so, we have defined a new CNN architecture using VGG16 convolutional part pre-trained on ImageNet and a custom fully connected part (Dense(128,'relu') connected to output Dense(2,'softmax')) trained with RGB images. 

\subsection{Training process}
To train the new CNN, we have downloaded the 25000 RGB images "Charm Myae Zaw and all" cat-dog dataset from Kaggle. The dataset contains 20000 images for training  and 5000 images for testing, each image size is 224x224x3.
The training program was written in Python using the Keras library and ImageDataGenerator(width\_shift\_range=0.1, height\_shift\_range=0.1, horizontal\_flip=True) for augmentation data. Due to hardware limitations, training only has 50 epochs but reaches an accuracy of 98\% over test data. 

\subsection{Contributions using R-LRP formula vs other LPR methods}
In a first step, we have calculated mask images (images of contributions) for all images from the test dataset, each mask image contains a percentage (1,5,10,15,20,25,40,50,60,75,80,85,90,95,99) of the highest contribution values over the three channels. Then, like for previous tests, we create, for each percentage, a dataset where images only contain the most relevant inputs, other inputs are set to 0.  

\subsection{Evaluations}
\begin{figure}[!h]  
  \centering
       \begin{tabular}{|>{\centering\arraybackslash}m{1.5cm}|>{\raggedright\arraybackslash}m{1.3cm}|>{\raggedright\arraybackslash}m{1.8cm}|>
       {\centering\arraybackslash}m{1.2cm}|>{\centering\arraybackslash}m{1.2cm}|>{\centering\arraybackslash}m{1.2cm}|>{\centering\arraybackslash}m{1.2cm}|>{\centering\arraybackslash}m{1.2cm}|}
        \hline
        Image& method & parameters & 1\% & 5\% & 10\% & 20\% & 50\%\\
        \hline
        \multirow{ 4}{*} {\includegraphics[width=1.4cm]{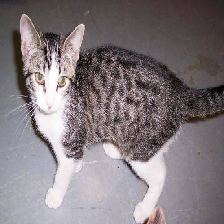} } &
         LRP-$\gamma$ & $\gamma=0.25$ &
        \includegraphics[width=1.2cm]{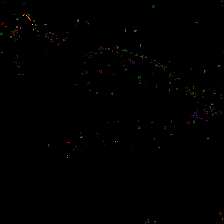} & 
        \includegraphics[width=1.2cm]{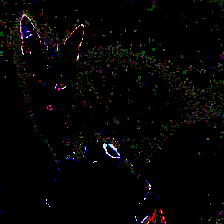}& 
        \includegraphics[width=1.2cm]{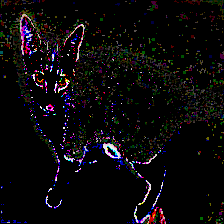} & \includegraphics[width=1.2cm]{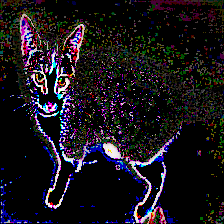}& \includegraphics[width=1.2cm]{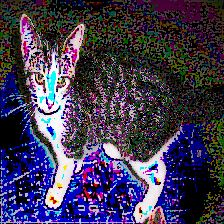} \\ \cline{2-7}
        & LRP-$\alpha\beta$ & $\alpha=2$ $\beta=-1$ & 
        \includegraphics[width=1.2cm]{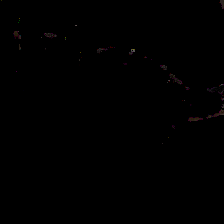} & 
        \includegraphics[width=1.2cm]{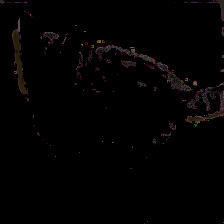}& 
        \includegraphics[width=1.2cm]{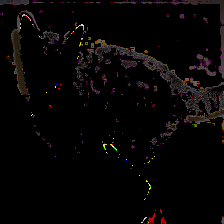} & \includegraphics[width=1.2cm]{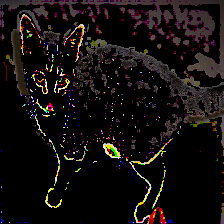}& \includegraphics[width=1.2cm]{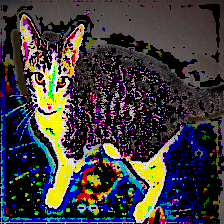}\\
        \cline{2-7}
        
        & R-LRP &  &
         \includegraphics[width=1.2cm]{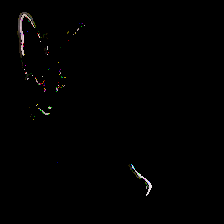} & 
        \includegraphics[width=1.2cm]{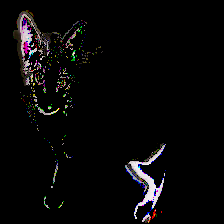}& 
        \includegraphics[width=1.2cm]{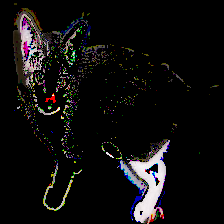} & \includegraphics[width=1.2cm]{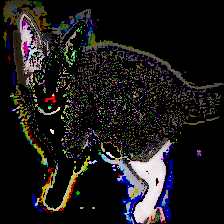}& \includegraphics[width=1.2cm]{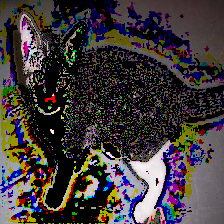}\\
        \hline
        \hline
        \multirow{ 4}{*} {\includegraphics[width=1.4cm]{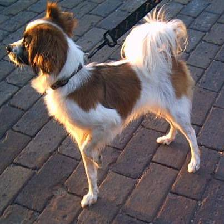} } &
         LRP-$\gamma$ & $\gamma=0.25$ &
        \includegraphics[width=1.2cm]{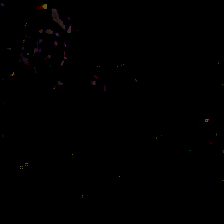} & 
        \includegraphics[width=1.2cm]{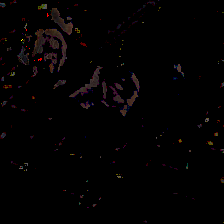}& 
        \includegraphics[width=1.2cm]{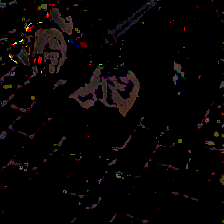} & \includegraphics[width=1.2cm]{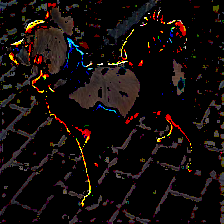}& \includegraphics[width=1.2cm]{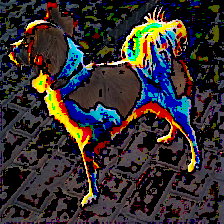} \\ \cline{2-7}
        & LRP-$\alpha\beta$ & $\alpha=2$ $\beta=-1$ & 
        \includegraphics[width=1.2cm]{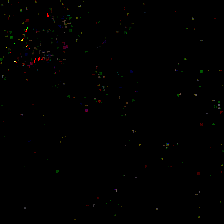} & 
        \includegraphics[width=1.2cm]{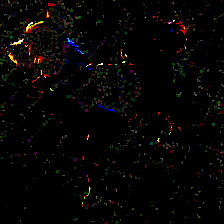}& 
        \includegraphics[width=1.2cm]{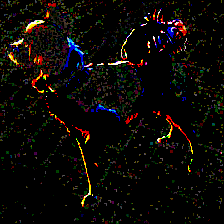} & \includegraphics[width=1.2cm]{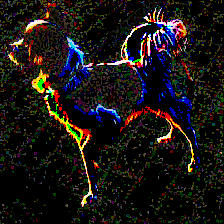}& \includegraphics[width=1.2cm]{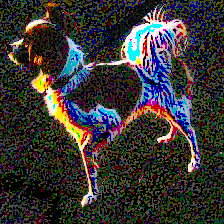}\\
        \cline{2-7}
        
        & R-LRP &  &
         \includegraphics[width=1.2cm]{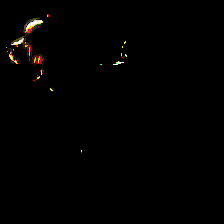} & 
        \includegraphics[width=1.2cm]{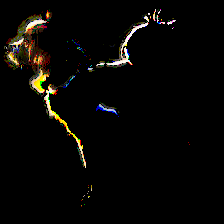}& 
        \includegraphics[width=1.2cm]{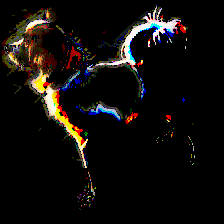} & \includegraphics[width=1.2cm]{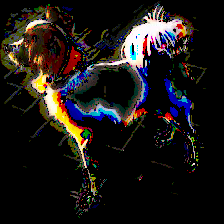}& \includegraphics[width=1.2cm]{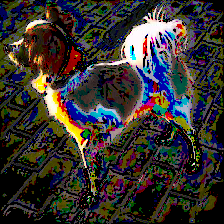}\\
        \hline
        \end{tabular}
        
  \caption{Two images, from the cat vs dog dataset, showing percentage of the most relevant inputs using some LRP methods with VGG16 based CNN network.}
    \label{fig:9}
\end{figure}

\begin{figure}[!h]  
  \centering
  \includegraphics[width=10cm]{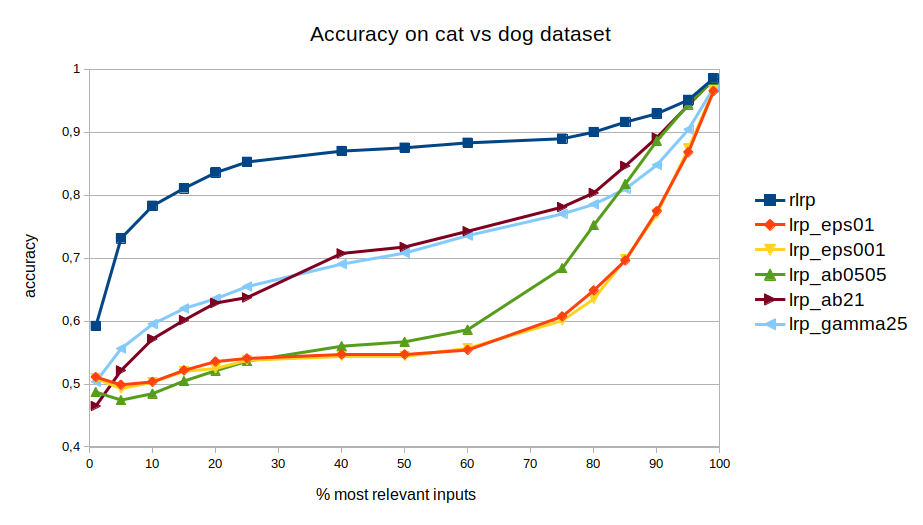}
    \caption{Accuracy of LRP methods over cat vs dog test dataset in terms of percentage of the most relevant inputs}
    \label{fig:10}
\end{figure}

As we can see in figure \ref{fig:9}, R-LRP avoids having non-relevant inputs in high contribution values. 
 Therefore, we have calculated the accuracy of the test dataset at each percentage using masked images in order to evaluate the LRP methods (figure \ref{fig:10}).

R-LRP outperforms all the other methods with an accuracy of 0.81 with only 15\% of the inputs of each image, whereas the over best score with the same percentage of inputs is 0.62.

\subsection{Evaluations using absolute value contributions}
\begin{figure}[!h]  
  \centering
       \begin{tabular}{|>{\centering\arraybackslash}m{1.5cm}|>{\raggedright\arraybackslash}m{1.3cm}|>{\raggedright\arraybackslash}m{1.8cm}|>
       {\centering\arraybackslash}m{1.2cm}|>{\centering\arraybackslash}m{1.2cm}|>{\centering\arraybackslash}m{1.2cm}|>{\centering\arraybackslash}m{1.2cm}|>{\centering\arraybackslash}m{1.2cm}|}
        \hline
        Image& method & parameters & 1\% & 5\% & 10\% & 20\% & 50\%\\
        \hline
        \multirow{ 4}{*} {\includegraphics[width=1.4cm]{images/data/cat_dog/init/image0cat_0_0_ori.png} } &
         LRP-$\gamma$ & $\gamma=0.25$ &
        \includegraphics[width=1.2cm]{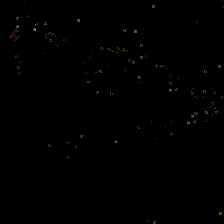} & 
        \includegraphics[width=1.2cm]{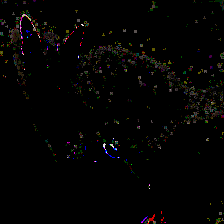}& 
        \includegraphics[width=1.2cm]{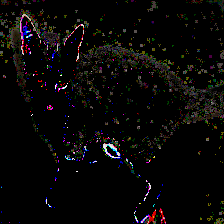} & \includegraphics[width=1.2cm]{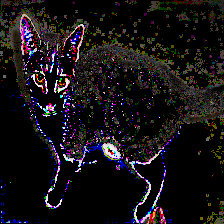}& \includegraphics[width=1.2cm]{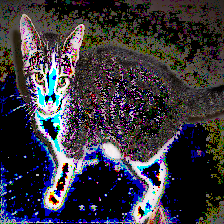} \\ \cline{2-7}
        & LRP-$\alpha\beta$ & $\alpha=2$ $\beta=-1$ & 
        \includegraphics[width=1.2cm]{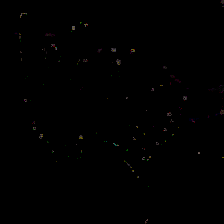} & 
        \includegraphics[width=1.2cm]{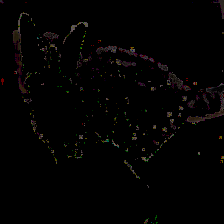}& 
        \includegraphics[width=1.2cm]{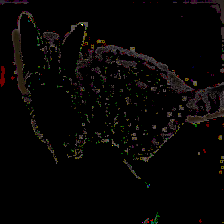} & \includegraphics[width=1.2cm]{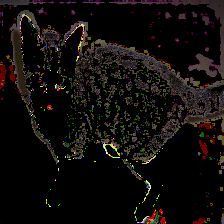}& \includegraphics[width=1.2cm]{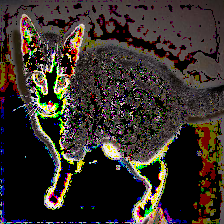}\\
        \cline{2-7}
        
        & R-LRP &  &
         \includegraphics[width=1.2cm]{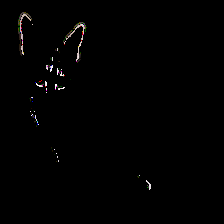} & 
        \includegraphics[width=1.2cm]{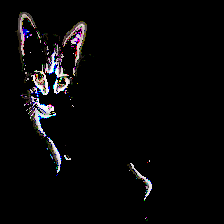}& 
        \includegraphics[width=1.2cm]{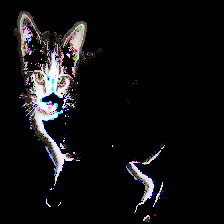} & \includegraphics[width=1.2cm]{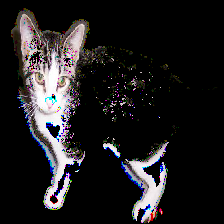}& \includegraphics[width=1.2cm]{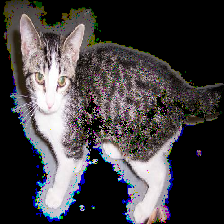}\\
        \hline
        \hline
        \multirow{ 4}{*} {\includegraphics[width=1.4cm]{images/data/cat_dog/init/image9cat_1_1_ori.png} } &
         LRP-$\gamma$ & $\gamma=0.25$ &
        \includegraphics[width=1.2cm]{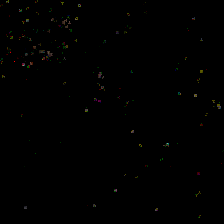} & 
        \includegraphics[width=1.2cm]{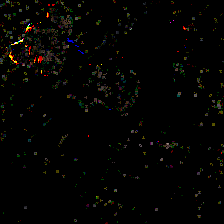}& 
        \includegraphics[width=1.2cm]{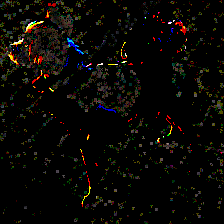} & \includegraphics[width=1.2cm]{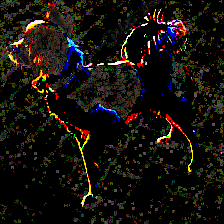}& \includegraphics[width=1.2cm]{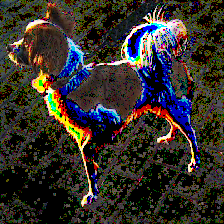} \\ \cline{2-7}
        & LRP-$\alpha\beta$ & $\alpha=2$ $\beta=-1$ & 
        \includegraphics[width=1.2cm]{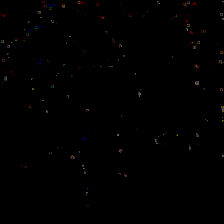} & 
        \includegraphics[width=1.2cm]{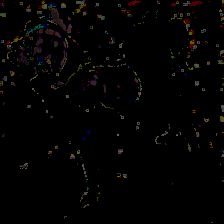}& 
        \includegraphics[width=1.2cm]{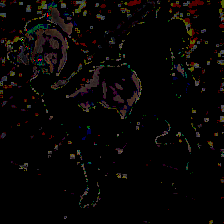} & \includegraphics[width=1.2cm]{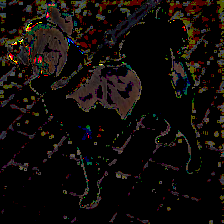}& \includegraphics[width=1.2cm]{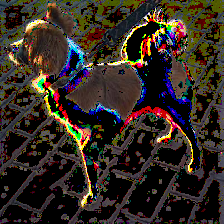}\\
        \cline{2-7}
        
        & R-LRP &  &
         \includegraphics[width=1.2cm]{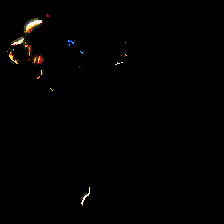} & 
        \includegraphics[width=1.2cm]{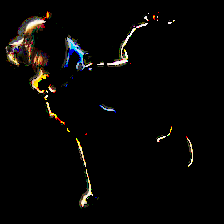}& 
        \includegraphics[width=1.2cm]{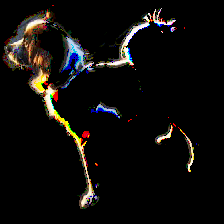} & \includegraphics[width=1.2cm]{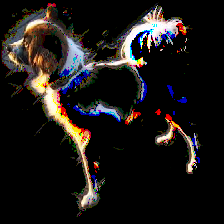}& \includegraphics[width=1.2cm]{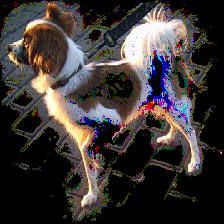}\\
        \hline
        \end{tabular}
        
  \caption{Two images, from the cat vs dog dataset, showing percentage of the most relevant inputs in absolute value using some LRP methods  with VGG16 based CNN network.}
    \label{fig:11}
\end{figure}

In figure \ref{fig:10}, we can note that, like for MNIST dataset, inputs with negative contribution values seem to be significant since after 80\% the scores increase sharply. 
So, we have calculated mask images that contain a percentage of the highest contributions in absolute value, like for previous evaluation (see Figure \ref{fig:11}).

As we can see in figure \ref{fig:11}, using absolute value for R-LRP contributions gives very good qualitative results, which means that inputs with the highest contributions are focused on (or near of) the object. This is what a human usually expect. There is a complementary between inputs with positive contributions and those with negative ones.

\begin{figure}[!h]  
  \centering
  \includegraphics[width=10cm]{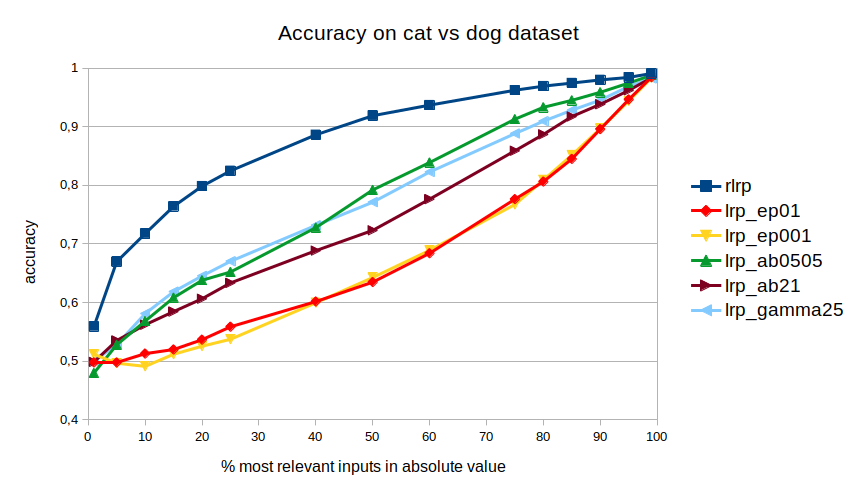}
    \caption{Accuracies of LRP methods over cat Vs dog test dataset in terms of percentage of the most relevant inputs in absolute value}
    \label{fig:12}
\end{figure}

For small percentage values, accuracy obtained with absolute value of contributions (figure \ref{fig:12}) are worse than in the previous evaluation (figure \ref{fig:10}) because the output for the identified category decrease due to negative contributions whereas the output for the other category is not affected in the same way. However, we can see that the R-LRP method gives better results than the other LRP methods.

\subsection{Evaluations with pixels}
\begin{figure}[!h]  
  \centering
       \begin{tabular}{|>{\centering\arraybackslash}m{1.5cm}|>{\raggedright\arraybackslash}m{1.3cm}|>{\raggedright\arraybackslash}m{1.8cm}|>
       {\centering\arraybackslash}m{1.2cm}|>{\centering\arraybackslash}m{1.2cm}|>{\centering\arraybackslash}m{1.2cm}|>{\centering\arraybackslash}m{1.2cm}|>{\centering\arraybackslash}m{1.2cm}|}
        \hline
        Image& method & parameters & 1\% & 5\% & 10\% & 20\% & 50\%\\
        \hline
        \multirow{ 4}{*} {\includegraphics[width=1.4cm]{images/data/cat_dog/init/image0cat_0_0_ori.png} } &
         LRP-$\gamma$ & $\gamma=0.25$ &
        \includegraphics[width=1.2cm]{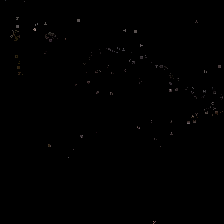} & 
        \includegraphics[width=1.2cm]{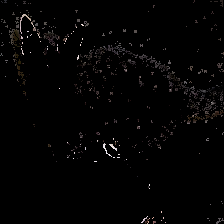}& 
        \includegraphics[width=1.2cm]{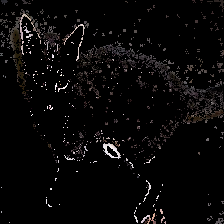} & \includegraphics[width=1.2cm]{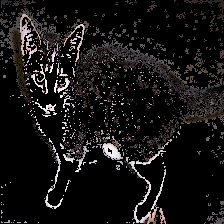}& \includegraphics[width=1.2cm]{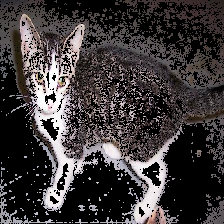} \\ \cline{2-7}
        & LRP-$\alpha\beta$ & $\alpha=2$ $\beta=-1$ & 
        \includegraphics[width=1.2cm]{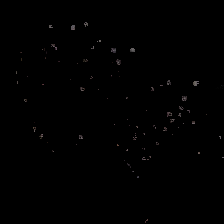} & 
        \includegraphics[width=1.2cm]{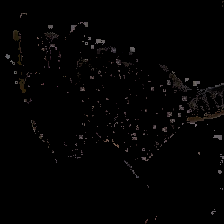}& 
        \includegraphics[width=1.2cm]{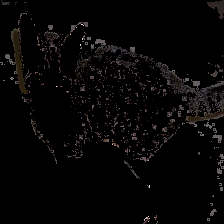} & \includegraphics[width=1.2cm]{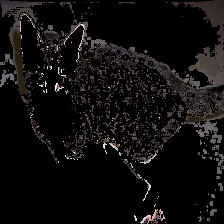}& \includegraphics[width=1.2cm]{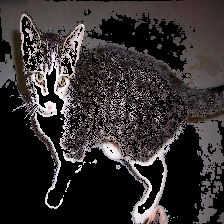}\\
        \cline{2-7}
        
        & R-LRP &  &
         \includegraphics[width=1.2cm]{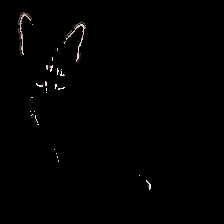} & 
        \includegraphics[width=1.2cm]{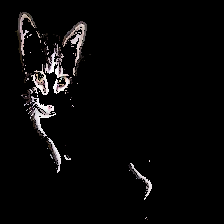}& 
        \includegraphics[width=1.2cm]{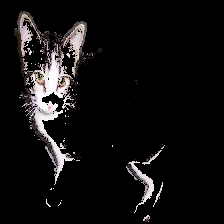} & \includegraphics[width=1.2cm]{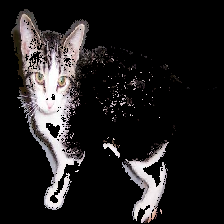}& \includegraphics[width=1.2cm]{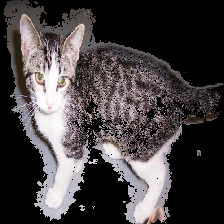}\\
        \hline
        \hline
        \multirow{ 4}{*} {\includegraphics[width=1.4cm]{images/data/cat_dog/init/image9cat_1_1_ori.png} } &
         LRP-$\gamma$ & $\gamma=0.25$ &
        \includegraphics[width=1.2cm]{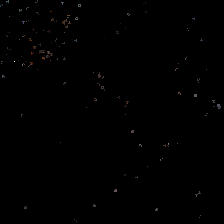} & 
        \includegraphics[width=1.2cm]{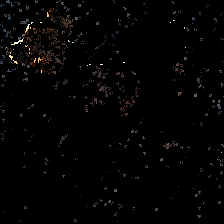}& 
        \includegraphics[width=1.2cm]{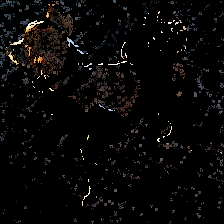} & \includegraphics[width=1.2cm]{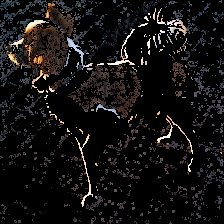}& \includegraphics[width=1.2cm]{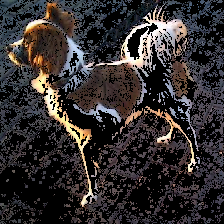} \\ \cline{2-7}
        & LRP-$\alpha\beta$ & $\alpha=2$ $\beta=-1$ & 
        \includegraphics[width=1.2cm]{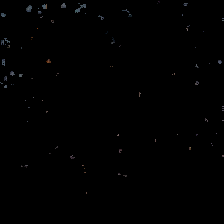} & 
        \includegraphics[width=1.2cm]{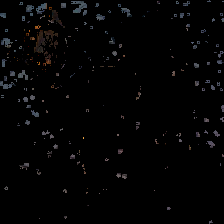}& 
        \includegraphics[width=1.2cm]{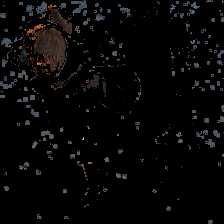} & \includegraphics[width=1.2cm]{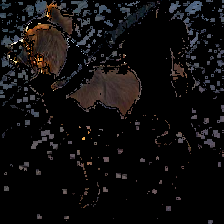}& \includegraphics[width=1.2cm]{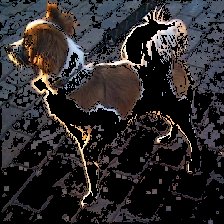}\\
        \cline{2-7}
        
        & R-LRP &  &
         \includegraphics[width=1.2cm]{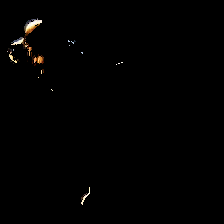} & 
        \includegraphics[width=1.2cm]{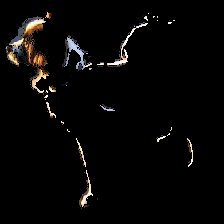}& 
        \includegraphics[width=1.2cm]{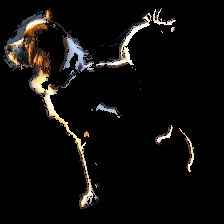} & \includegraphics[width=1.2cm]{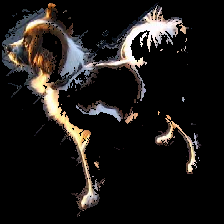}& \includegraphics[width=1.2cm]{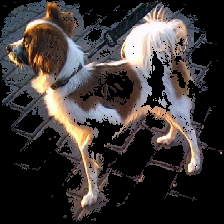}\\
        \hline
        \end{tabular}
  \caption{Two images, from the cat Vs dog dataset, showing percentage of the most relevant pixels in absolute value using some LRP methods  with VGG16 based CNN network.}
    \label{fig:11p}
\end{figure}

In the previous tests, we have used input values, which means that, for one pixel, we masked separately each input channel. Another point of view is to consider the pixel entity in the input image and thus, the contribution value for a pixel is the direct sum of the contribution over each channel. We present tests using absolute values of these pixels contributions, which give the best result in terms of quality (see figure \ref{fig:11p}).

\begin{figure}[!h]  
  \centering
  \includegraphics[width=10cm]{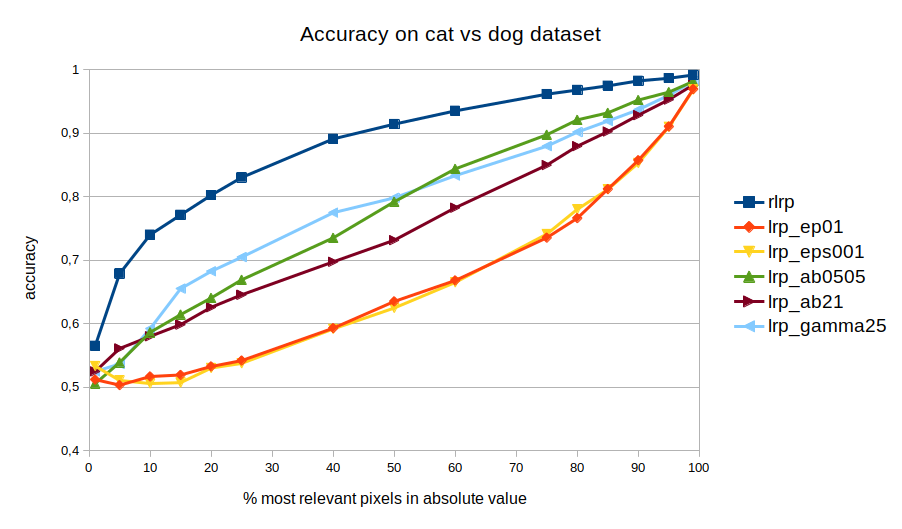}
    \caption{Accuracy of LRP methods over cat Vs dog test dataset in terms of percentage of the most relevant pixels in absolute value}
    \label{fig:12p}
\end{figure}

As expected, the test using most relevant pixel contributions (in absolute value) gives similar results (figure \ref{fig:12p}) than test using most relevant input contributions (figure \ref{fig:12}) and it also shows that our R-LRP method gives better results than other LRP ones for small percentage values of relevant pixels.

To complete our method's validation, we have used it over more complex CNN networks.

\section{Evaluation on Imagenet1K dataset with VGG16, VGG19 and ResNet50}

We present the contribution results obtained on the "Synthetic ImageNet-1K" dataset available on Kaggle. We used pretrained Keras VGG16, VGG19 and ResNet50 networks on this dataset without any fine-tuning. For computing reasons, we used only two random images for each category from the dataset. We keep the same images for all our tests. For an image, the category taken into account for the tests is the highest output of the network we use, as some categories do not match between Imagenet and Synthetic ImageNet-1K datasets. 
Only pixels contributions (sum of RGB contributions) are presented in this part. As for previous evaluations, we have produced one dataset for each method and each percentage of most relevant pixels in absolute value and used them for accuracy evaluation.

\subsection{VGG16 and VGG19}
\begin{figure}[!h]  
  \centering
       \begin{tabular}{|>{\centering\arraybackslash}m{1.5cm}|>{\raggedright\arraybackslash}m{1.3cm}|>{\raggedright\arraybackslash}m{1.8cm}|>
       {\centering\arraybackslash}m{1.2cm}|>{\centering\arraybackslash}m{1.2cm}|>{\centering\arraybackslash}m{1.2cm}|>{\centering\arraybackslash}m{1.2cm}|>{\centering\arraybackslash}m{1.2cm}|}
        \hline
        Image& method &parameters & 1\% & 5\% & 10\% & 20\% & 50\%\\
        \hline
        \multirow{ 4}{*} {\includegraphics[width=1.4cm]{images/data/imagenet/init/lot0_image2cat_1_1_ori.png} } &
         LRP-$\gamma$ & $\gamma=0.25$ &
        \includegraphics[width=1.2cm]{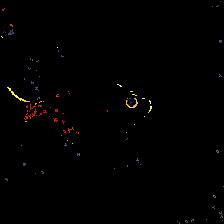} & 
        \includegraphics[width=1.2cm]{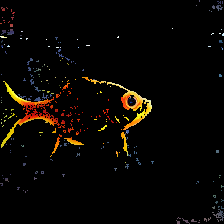}& 
        \includegraphics[width=1.2cm]{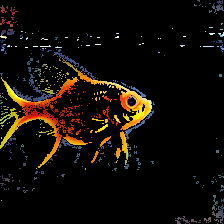} & \includegraphics[width=1.2cm]{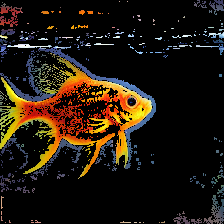}& \includegraphics[width=1.2cm]{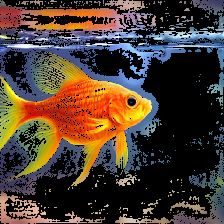} \\ \cline{2-7}
        & LRP-$\alpha\beta$ & $\alpha=2$ $\beta=-1$ & 
        \includegraphics[width=1.2cm]{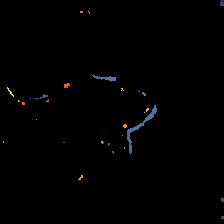} & 
        \includegraphics[width=1.2cm]{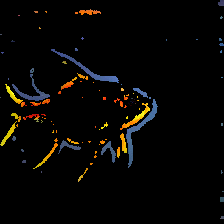}& 
        \includegraphics[width=1.2cm]{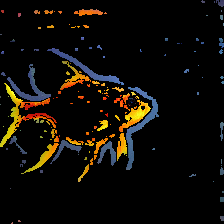} & \includegraphics[width=1.2cm]{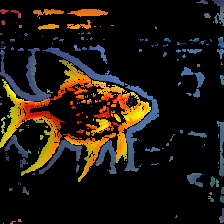}& \includegraphics[width=1.2cm]{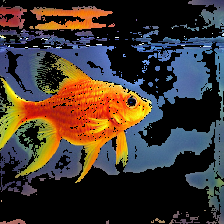}\\
        \cline{2-7}
        
        & R-LRP &  &
         \includegraphics[width=1.2cm]{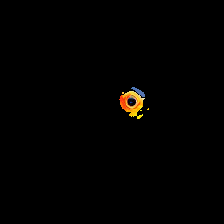} & 
        \includegraphics[width=1.2cm]{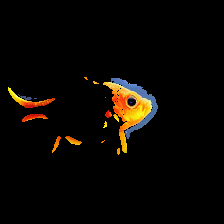}& 
        \includegraphics[width=1.2cm]{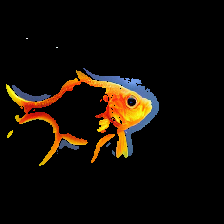} & \includegraphics[width=1.2cm]{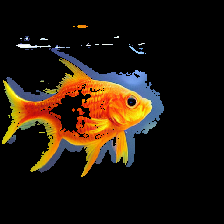}& \includegraphics[width=1.2cm]{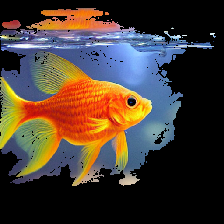}\\
        \hline
        \hline
        \multirow{ 4}{*} {\includegraphics[width=1.4cm]{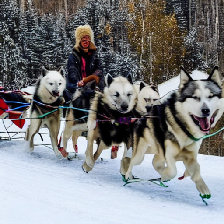} } &
         LRP-$\gamma$ & $\gamma=0.25$ &
        \includegraphics[width=1.2cm]{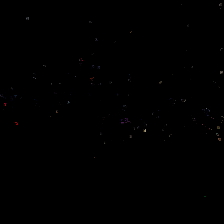} & 
        \includegraphics[width=1.2cm]{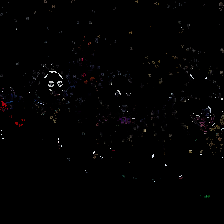}& 
        \includegraphics[width=1.2cm]{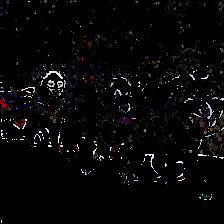} & \includegraphics[width=1.2cm]{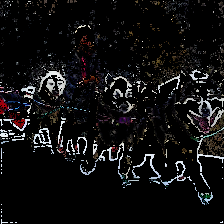}& \includegraphics[width=1.2cm]{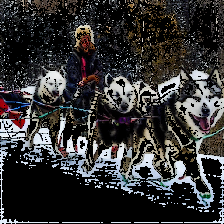} \\ \cline{2-7}
        & LRP-$\alpha\beta$ & $\alpha=2$ $\beta=-1$ & 
        \includegraphics[width=1.2cm]{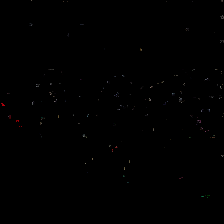} & 
        \includegraphics[width=1.2cm]{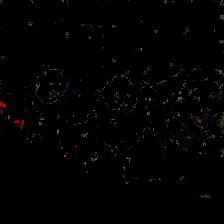}& 
        \includegraphics[width=1.2cm]{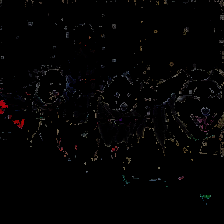} & \includegraphics[width=1.2cm]{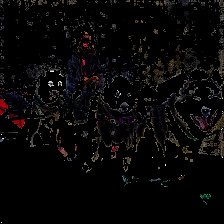}& \includegraphics[width=1.2cm]{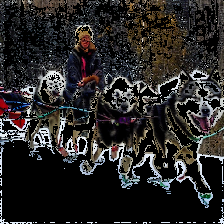}\\
        \cline{2-7}
        & R-LRP &  &
         \includegraphics[width=1.2cm]{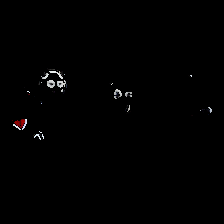} & 
        \includegraphics[width=1.2cm]{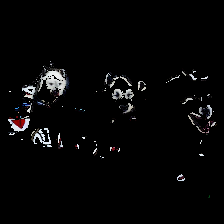}& 
        \includegraphics[width=1.2cm]{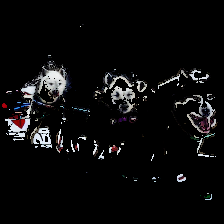} & \includegraphics[width=1.2cm]{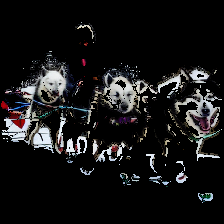}& \includegraphics[width=1.2cm]{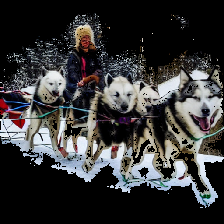}\\
        \hline
        \hline
        \multirow{ 4}{*} {\includegraphics[width=1.4cm]{images/data/imagenet/init/lot56_image2cat_561_561_ori.png} } &
         LRP-$\gamma$ & $\gamma=0.25$ &
        \includegraphics[width=1.2cm]{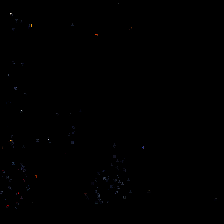} & 
        \includegraphics[width=1.2cm]{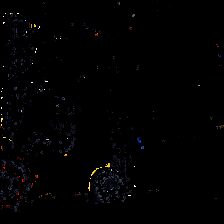}& 
        \includegraphics[width=1.2cm]{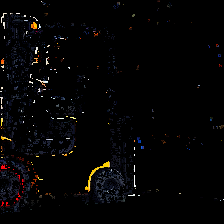} & \includegraphics[width=1.2cm]{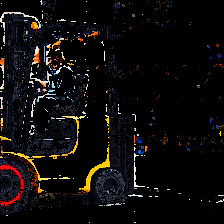}& \includegraphics[width=1.2cm]{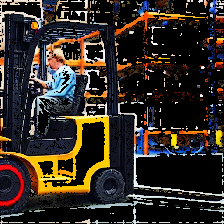} \\ \cline{2-7}
        & LRP-$\alpha\beta$ & $\alpha=2$ $\beta=-1$ & 
        \includegraphics[width=1.2cm]{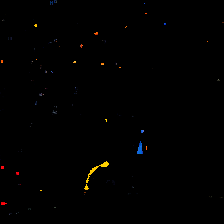} & 
        \includegraphics[width=1.2cm]{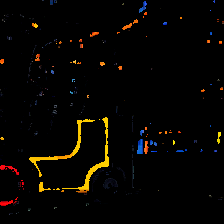}& 
        \includegraphics[width=1.2cm]{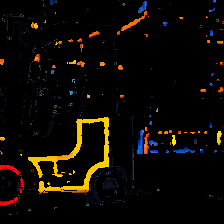} & \includegraphics[width=1.2cm]{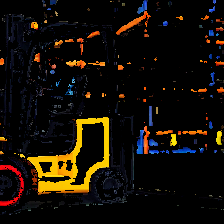}& \includegraphics[width=1.2cm]{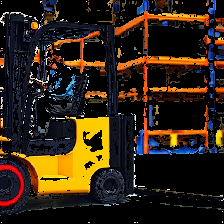}\\
        \cline{2-7}
        & R-LRP &  &
         \includegraphics[width=1.2cm]{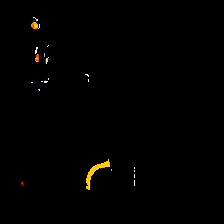} & 
        \includegraphics[width=1.2cm]{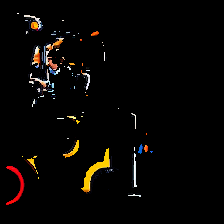}& 
        \includegraphics[width=1.2cm]{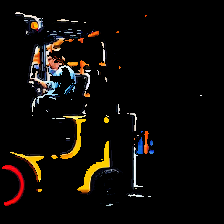} & \includegraphics[width=1.2cm]{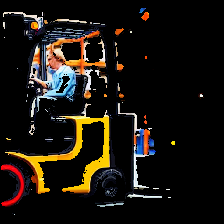}& \includegraphics[width=1.2cm]{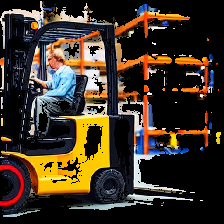}\\
        \hline
        \end{tabular}
        
  \caption{percentage of the most relevant pixels in absolute value using some LRP methods with VGG16 network on some challenging images from Synthetic ImageNet-1K dataset.}
    \label{fig:13_pix}
\end{figure}

\begin{figure}[!h]  
  \centering
  \begin{tabular}{c c}
  \includegraphics[width=7cm]{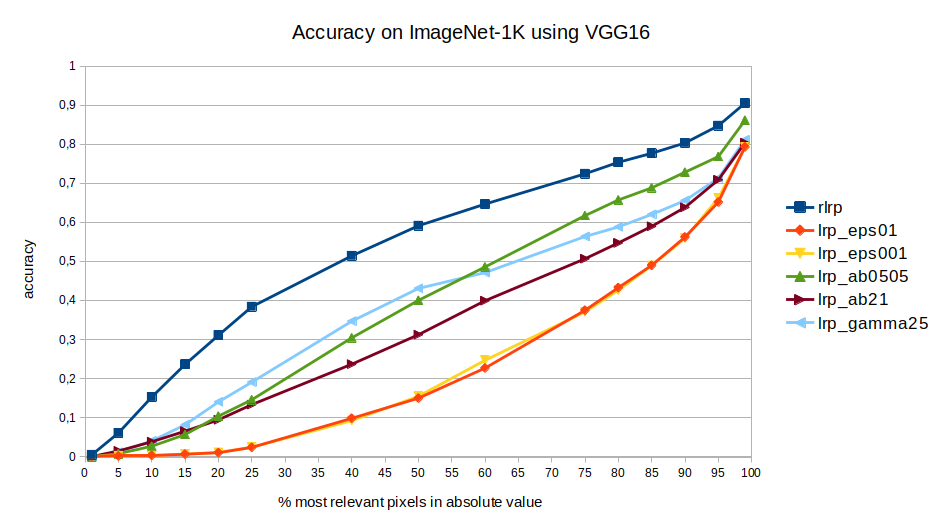}
  &
  \includegraphics[width=7cm]{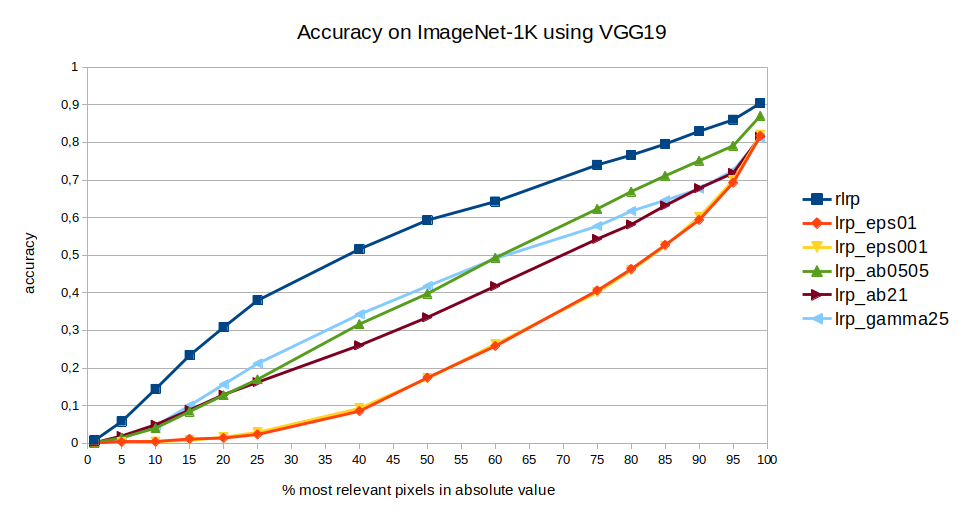}
  \end{tabular}
    \caption{Accuracy of LRP methods in terms of percentage of the most relevant pixels in absolute value for VGG16 and VGG19 networks with Synthetic ImageNet-1K}
    \label{fig:14}
\end{figure}

Qualitatively, R-LRP gives much more inputs focused on objects than other LRP methods at the different percentage of most relevant values in absolute value (see figure \ref{fig:13_pix}). 
Moreover, accuracy for these different percentage is better with our method than any other LRP method (see figure \ref{fig:14}). Because of the 1000 classification categories, the accuracy is very low with only 1\% of pixels (501 pixels) as we have a little more than one chance in a thousand of finding the right category. As VGG16 and VGG19 achieve a test accuracy of around 90\% on ImageNet (source https://keras.io/api/applications/), we assume that on ImageNet-1K, we get an accuracy of 90\%, which means that in figure \ref{fig:14} the accuracies with 99\% of most relevant pixels only reach 0.90. LRP-$\epsilon$ method gives the worst results with an accuracy around 2.5\% with 25\% of relevant pixels in images, so we focused on other methods.

Because VGG16 and VGG19 networks are not so different, the test results are similar. 

\subsection{ResNet50}
When we try to evaluate contributions using LRP-$\epsilon$ with ResNet50, overflow occurs due to accumulation of division by small values. To overcome the overflow, we have to use $\epsilon>1$, which is too big and gives bad contributions. For LRP-$\gamma$, we got also very bad contribution images.
As LRP-$\epsilon$ and LRP-$\gamma$ methods are not suitable to evaluate contributions with ResNet50, we have only compared our R-LRP method with LRP-$\alpha\beta$ one. For further examples, you can take a look at the article \cite{Jee2021} which, however, uses heatmaps.

\begin{figure}[!h]  
  \centering
       \begin{tabular}{|>{\centering\arraybackslash}m{1.5cm}|>{\raggedright\arraybackslash}m{1.3cm}|>{\raggedright\arraybackslash}m{1.8cm}|>
       {\centering\arraybackslash}m{1.2cm}|>{\centering\arraybackslash}m{1.2cm}|>{\centering\arraybackslash}m{1.2cm}|>{\centering\arraybackslash}m{1.2cm}|>{\centering\arraybackslash}m{1.2cm}|}
        \hline
        Image& method & parameters & 1\% & 5\% & 10\% & 20\% & 50\%\\
        \hline
        \multirow{ 4}{*} {\includegraphics[width=1.4cm]{images/data/imagenet/init/lot0_image2cat_1_1_ori.png} } &
         LRP-$\alpha\beta$ & $\alpha=0.5$ $\beta=0.5$ &
        \includegraphics[width=1.2cm]{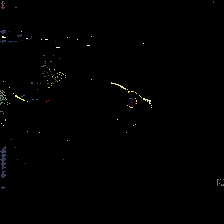} & 
        \includegraphics[width=1.2cm]{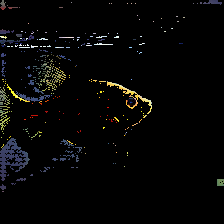}& 
        \includegraphics[width=1.2cm]{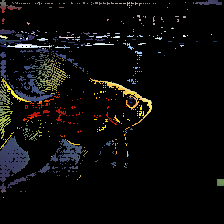} & \includegraphics[width=1.2cm]{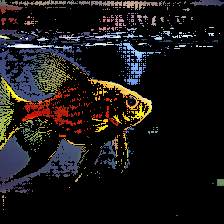}& \includegraphics[width=1.2cm]{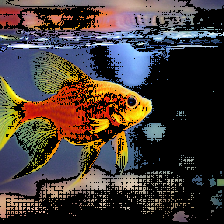} \\ \cline{2-7}
        & LRP-$\alpha\beta$ & $\alpha=2$ $\beta=-1$ & 
        \includegraphics[width=1.2cm]{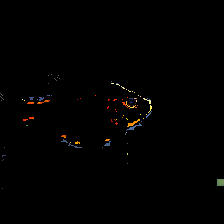} & 
        \includegraphics[width=1.2cm]{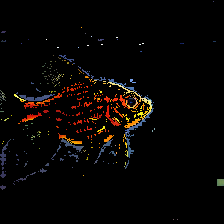}& 
        \includegraphics[width=1.2cm]{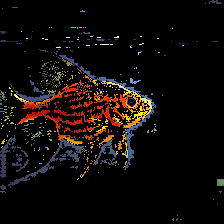} & \includegraphics[width=1.2cm]{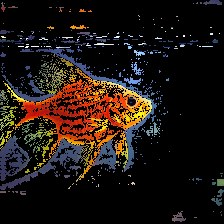}& \includegraphics[width=1.2cm]{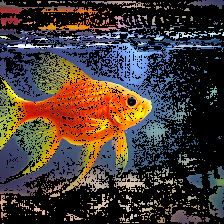}\\
        \cline{2-7}
        
        & R-LRP &  &
         \includegraphics[width=1.2cm]{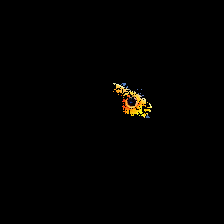} & 
        \includegraphics[width=1.2cm]{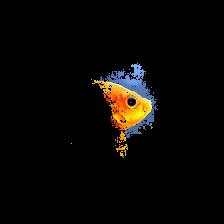}& 
        \includegraphics[width=1.2cm]{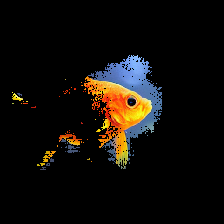} & \includegraphics[width=1.2cm]{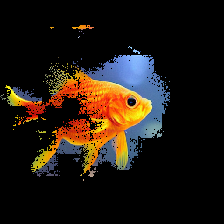}& \includegraphics[width=1.2cm]{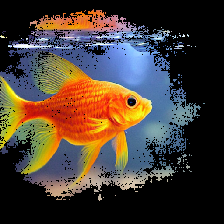}\\
        \hline
        \hline
        \multirow{ 4}{*} {\includegraphics[width=1.4cm]{images/data/imagenet/init/lot53_image14cat_537_537_ori.png} } &
         LRP-$\alpha\beta$ & $\alpha=0.5$ $\beta=0.5$ &
        \includegraphics[width=1.2cm]{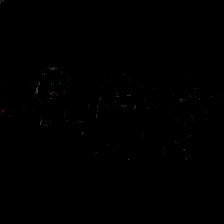} & 
        \includegraphics[width=1.2cm]{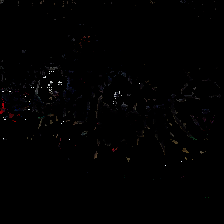}& 
        \includegraphics[width=1.2cm]{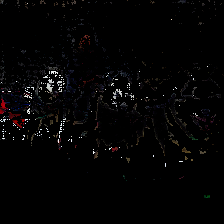} & \includegraphics[width=1.2cm]{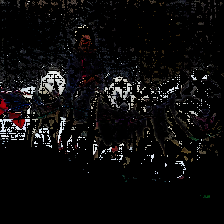}& \includegraphics[width=1.2cm]{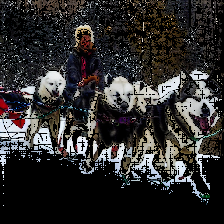} \\ \cline{2-7}
        & LRP-$\alpha\beta$ & $\alpha=2$ $\beta=-1$ & 
        \includegraphics[width=1.2cm]{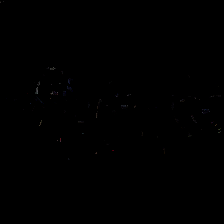} & 
        \includegraphics[width=1.2cm]{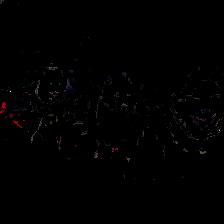}& 
        \includegraphics[width=1.2cm]{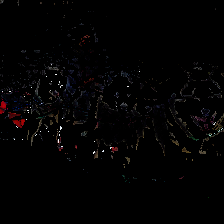} & \includegraphics[width=1.2cm]{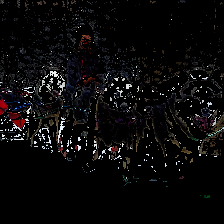}& \includegraphics[width=1.2cm]{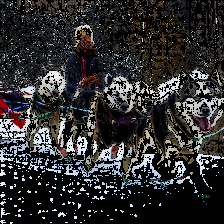}\\
        \cline{2-7}
        & R-LRP &  &
         \includegraphics[width=1.2cm]{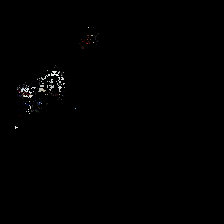} & 
        \includegraphics[width=1.2cm]{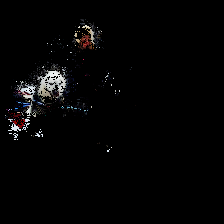}& 
        \includegraphics[width=1.2cm]{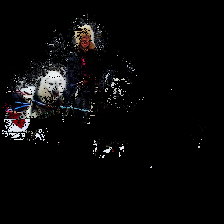} & \includegraphics[width=1.2cm]{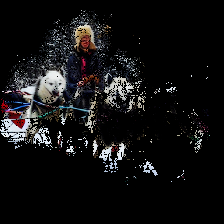}& \includegraphics[width=1.2cm]{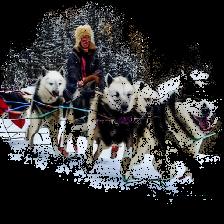}\\
        \hline
        \hline
        \multirow{ 4}{*} {\includegraphics[width=1.4cm]{images/data/imagenet/init/lot56_image2cat_561_561_ori.png} } &
         LRP-$\alpha\beta$ & $\alpha=0.5$ $\beta=0.5$ &
        \includegraphics[width=1.2cm]{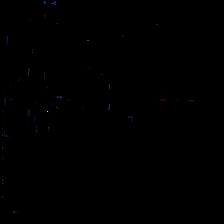} & 
        \includegraphics[width=1.2cm]{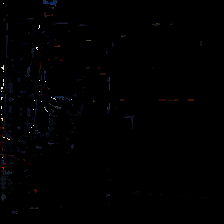}& 
        \includegraphics[width=1.2cm]{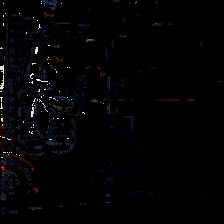} & \includegraphics[width=1.2cm]{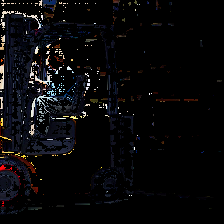}& \includegraphics[width=1.2cm]{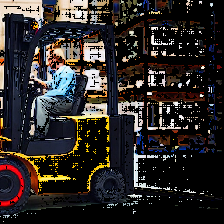} \\ \cline{2-7}
        & LRP-$\alpha\beta$ & $\alpha=2$ $\beta=-1$ & 
        \includegraphics[width=1.2cm]{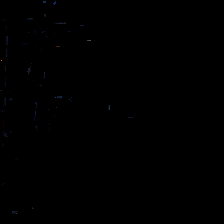} & 
        \includegraphics[width=1.2cm]{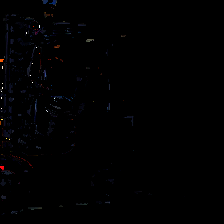}& 
        \includegraphics[width=1.2cm]{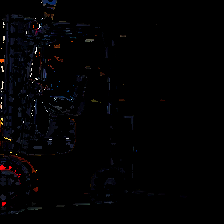} & \includegraphics[width=1.2cm]{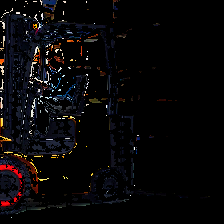}& \includegraphics[width=1.2cm]{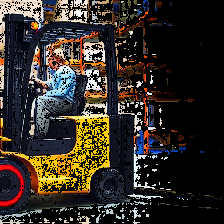}\\
        \cline{2-7}
        & R-LRP &  &
         \includegraphics[width=1.2cm]{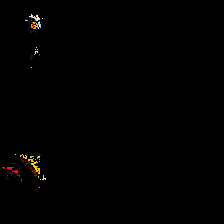} & 
        \includegraphics[width=1.2cm]{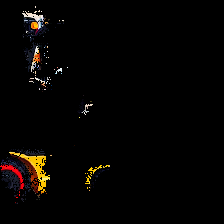}& 
        \includegraphics[width=1.2cm]{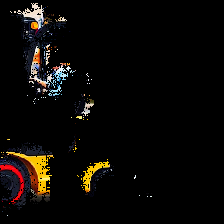} & \includegraphics[width=1.2cm]{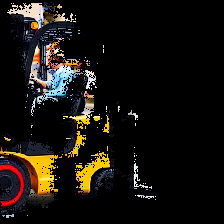}& \includegraphics[width=1.2cm]{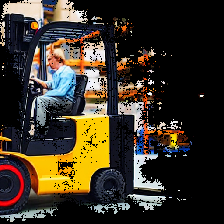}\\
        \hline
        \end{tabular}
  \caption{percentage of the most relevant pixels in absolute value using some LRP methods with ResNet50 network on some challenging images from Synthetic ImageNet-1K dataset.}
    \label{fig:resnet50_pix}
\end{figure}
\begin{figure}[!h]  
  \centering
  \includegraphics[width=8cm]{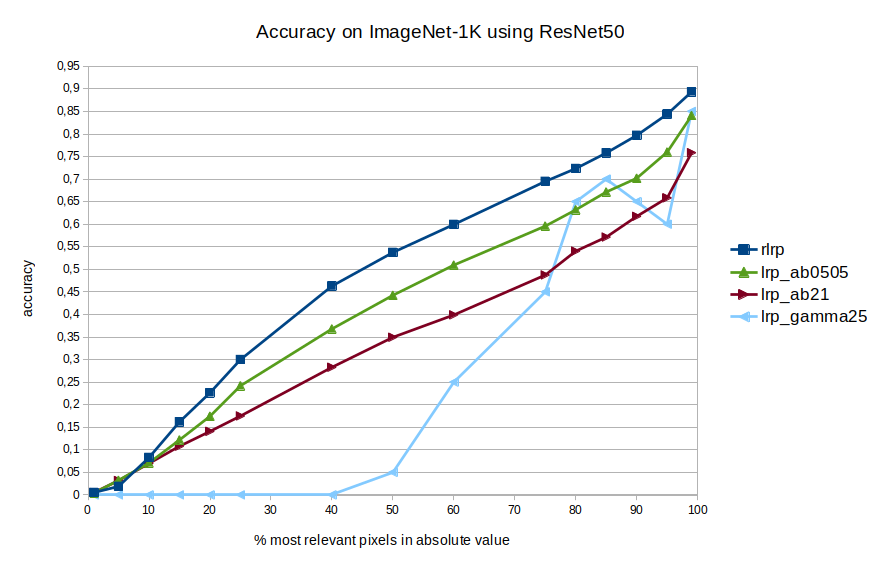}
    \caption{Accuracy of LRP methods over ImageNet-1K dataset in terms of percentage of the most relevant pixels in absolute value using ResNet50}
    \label{fig:resnet50_accu}
\end{figure}

We can see in figure \ref{fig:resnet50_pix} that our method gives better results than LRP-$\alpha\beta$ with ResNet50 as relevant pixels are more focused on objects with R-LRP. Furthermore, using the same test to evaluate the accuracy of imagenet-1K image contributions in terms of percentage of most relevant pixels (figure \ref{fig:resnet50_accu}), our method gives better results. Like for VGG evaluations, accuracy goes from 0.001 to 0.91 as ResNet50 pretrained model from KERAS achieves 92\% of accuracy.

\subsection{Decision process comparison}
\begin{figure}[!h]  
  \centering
  \begin{tabular}{c c c}
  \includegraphics[width=4.3cm]{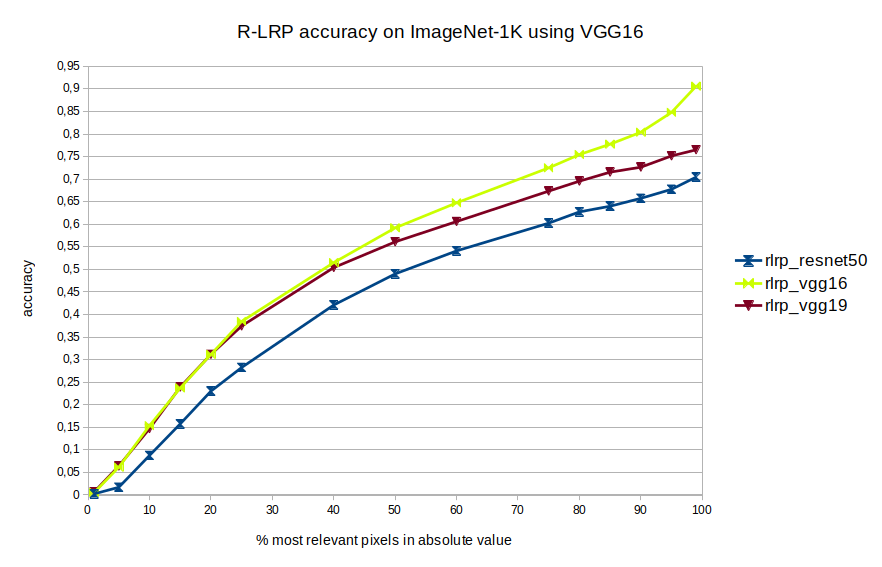}
  &
  \includegraphics[width=4.3cm]{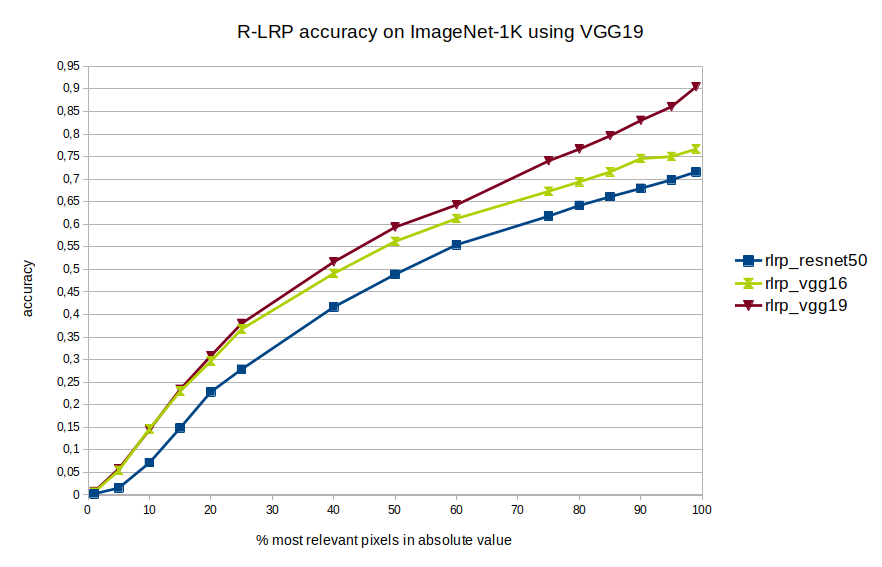}
  &
  \includegraphics[width=4.3cm]{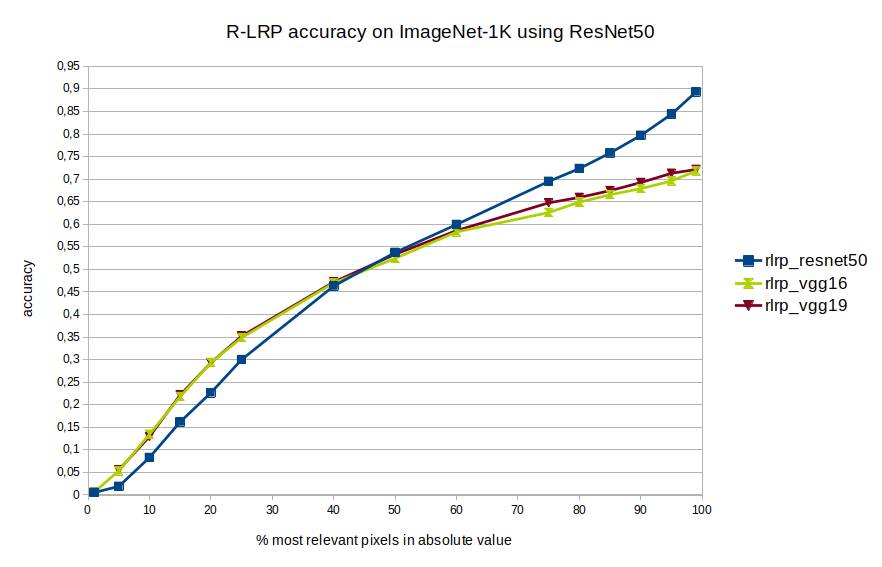}\\
  (a) & (b) & (c)
  \end{tabular}
    \caption{Cross comparison of networks using R-LRP. Relevant pixels for one network on ImageNet-1K are used to evaluate accuracy with another one (from left to right, VGG16, VGG19 and ResNet50)}
    \label{fig:cross_rlrp}
\end{figure}

We want to find out if the networks use the same pixels in order to make their decision. As R-LRP gives for each image and for each network the relevant pixels, we use relevant pixels for one network in order to evaluate the accuracy using another network. In figure \ref{fig:cross_rlrp}.a and figure \ref{fig:cross_rlrp}.b, we can see that VGG19 and VGG16 use, in general, the same pixels to class images. On the other hand (see Figure \ref{fig:cross_rlrp}.c), we can show that VGG networks and ResNet50 use different pixels in order to identify images. Indeed, utilizing VGG networks yields an accuracy of 82\%, despite having 99\% of the most relevant pixels from ResNet50 for each image. Moreover, Resnet seems to be more "conceptual" using blobs/areas whereas VGG networks rely on protruding or curved shapes.

Evaluating the 2000 images used for the tests with the three networks, it shows that 1464 images get the same category with VGG16 and VGG19, 1178 images get the same category with VGG16 and ResNet50 and 1194 images get the same category with VGG19 and ResNet50. This confirms the charts in figure \ref{fig:cross_rlrp}.

\section{Qualitative evaluation of LRP methods}
In order to evaluate the quality of the LRP methods, we downloaded the "explainable ai imagenet 12" dataset from Kaggle. The primary goal of this dataset is object recognition, with assistance from the background. The pictures in this set are in 12 folders with 1300 pictures in each folder. Although images in the same folder contain similar objects, several categories are available for images in one folder and no category have been associated with each folder. Thus, to be usable for qualitative evaluations of LRP methods, we only kept images of the most represented category in each folder. Categories have been defined using Keras VGG16 pretrained network on ImageNet. Since this dataset also contains object masks in images, we have access to the positions of the objects in the images. However, some object's mask images are not good for the evaluations (almost completely black or white images) and we only retained images in which the corresponding object's mask represents between 1\% and 99\% of the image. In the end, we obtained 12 folders  of images corresponding to 12 categories (44, 46, 124, 147, 153, 277, 393, 405, 479, 480, 486 and 928), each folder contains between 680 and 1140 images. Contrary to quantitative evaluations, qualitative evaluations have been done with a limited number of categories and with imperfect object mask images. So the results of this part just provides indicative elements on some categories.

Like for previous tests, we have created for all images in all folders, for each percentage and for each method, mask images which only contain the most relevant pixels set to white and the other pixels set to black. As folders (categories) do not contain the same amount of images, the tests have been done separately for each category.

The first measure we have done, is the percentage of relevant pixels in the object mask also named the pointing game in \cite{Zhang2018}. The second measure is the average distance between the relevant pixels and the object mask.

\subsection{Mask pointing game}
\begin{figure}[!h]  
  \centering
       \begin{tabular}{|>{\centering\arraybackslash}m{1.8cm}|>{\centering\arraybackslash}m{1.8cm}|>{\centering\arraybackslash}m{1.8cm}|>{\centering\arraybackslash}m{2.1cm}|>{\centering\arraybackslash}m{1.8cm}|>{\centering\arraybackslash}m{2.1cm}|}
        \hline
        Image& Mask & R-LRP & LRP\_ab0505 & R-LRP & LRP\_ab0505 \\
        & & VGG16 & VGG16 & ResNet50 & ResNet50 \\
        \hline
        \includegraphics[width=1.4cm]{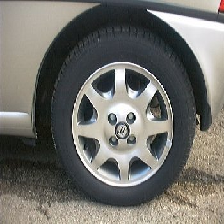}  &
        \includegraphics[width=1.4cm]{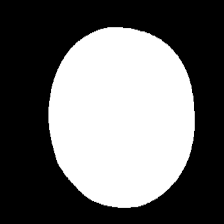} & 
        \includegraphics[width=1.4cm]{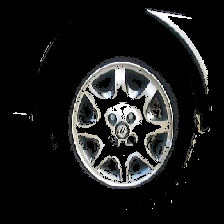}& 
        \includegraphics[width=1.4cm]{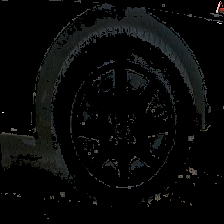} & 
        \includegraphics[width=1.4cm]{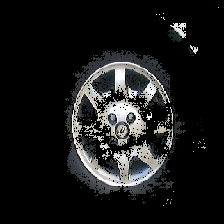}& 
        \includegraphics[width=1.4cm]{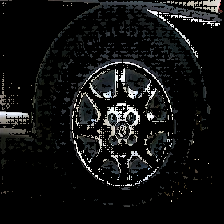} \\
        \hline
        \multicolumn{2}{|c|}{Average pixels in mask} & 0.8662 & 0.3525 & 0.9687 & 0.5808 \\
        
        \hline
        \end{tabular}
        
  \caption{Mask and images with 20$\%$ of most relevant pixels using R-LRP and LRP-ab0505 for image 1040 (category 479)}
    \label{fig:1040_pix}
\end{figure}

First, we calculate the proportion of relevant pixels within object masks (see Figure \ref{fig:1040_pix}).
By averaging the values over all images in a category, we get a score of the method for that category. The higher the score for a small percentage of pixels, the better is the method (in a human point of view). However, the larger the mask, the more likely a pixel is to be inside. So the results obtained must be put into perspective.
\begin{figure}[!h]  
  \centering
  \begin{tabular}{c c}
  \includegraphics[width=6cm]{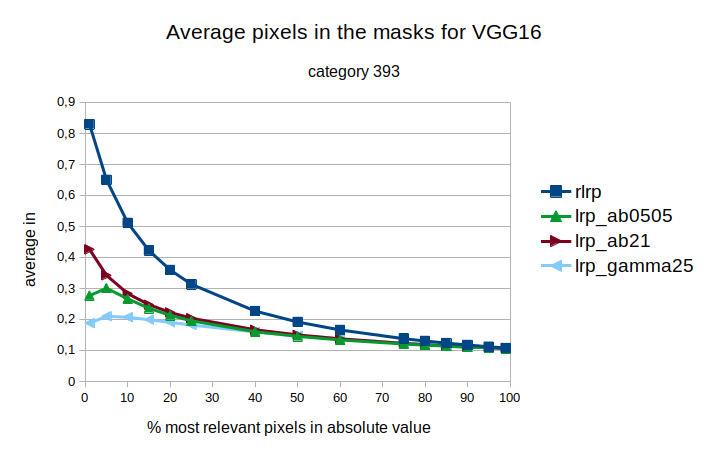}
  &
  \includegraphics[width=6cm]{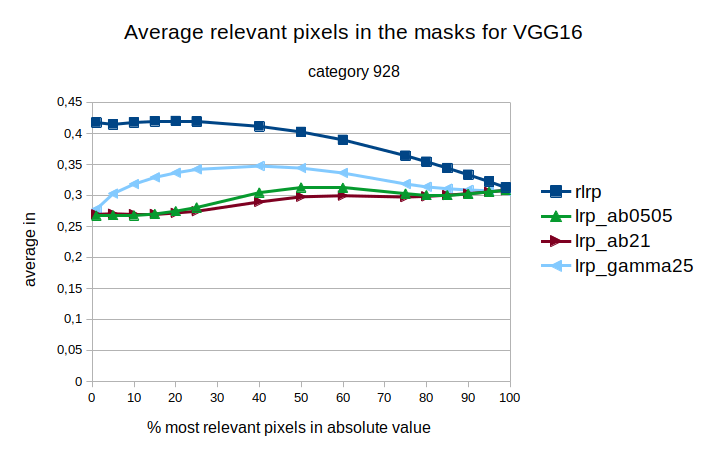}
  \end{tabular}
    \caption{"Best" (cat 393) and "worst" (cat 928) average pixels in the mask  for VGG16}
    \label{fig:b_w_in_vgg16}
\end{figure}
In figure \ref{fig:b_w_in_vgg16}, we can see that, for the category 393 ('anemone fish'), R-LRP gives a very good score. This mean that, for that category, R-LRP locates very well the objects in the images according to the defined masks. For category 928 ('ice cream'), none of the LRP methods give good scores : in fact, it seems that, in general,  the relevance of the pixels doesn't correspond to masks and so the source of decision of the network is not that human expects. Evaluations with ResNet50 (see figure \ref{fig:b_w_in_resnet}) confirm that, for category 928, all LRP methods give poor results. Which seems to confirm that the masks and the relevant pixels do not match for that category.
\begin{figure}[!h]  
  \centering
  \begin{tabular}{c c}
  \includegraphics[width=6cm]{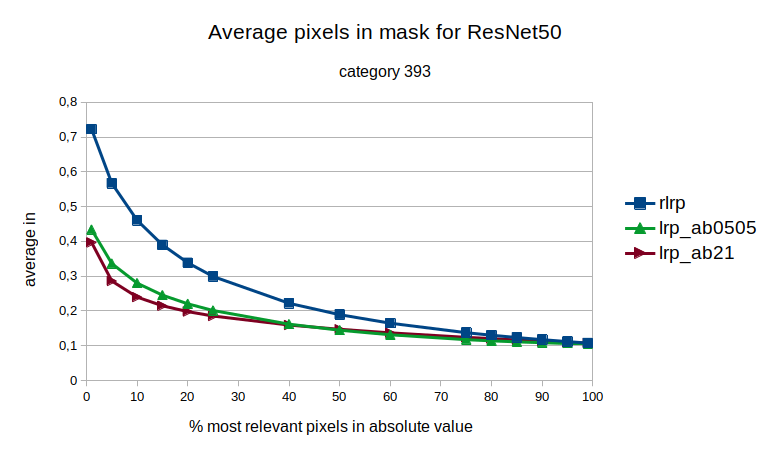}
  &
  \includegraphics[width=6cm]{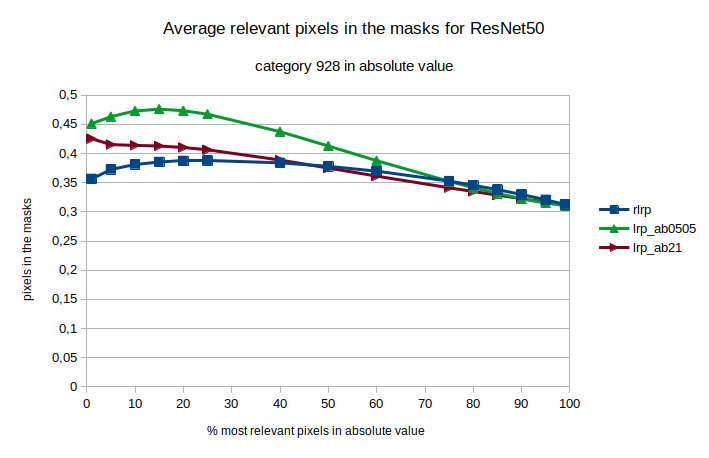}
  \end{tabular}
    \caption{"Best" (cat 393) and "worst" (cat 928) average pixels in the mask for ResNet50}
    \label{fig:b_w_in_resnet}
\end{figure}

\subsection{Average distances}
\begin{figure}[!h]  
  \centering
       \begin{tabular}{|>{\centering\arraybackslash}m{1.8cm}|>{\centering\arraybackslash}m{1.8cm}|>{\centering\arraybackslash}m{1.8cm}|>{\centering\arraybackslash}m{2.1cm}|>{\centering\arraybackslash}m{1.8cm}|>{\centering\arraybackslash}m{2.1cm}|}
        \hline
        Image& Mask & R-LRP & LRP\_ab0505 & R-LRP & LRP\_ab0505 \\
        & & VGG16 & VGG16 & ResNet50 & ResNet50 \\
        \hline
        \includegraphics[width=1.4cm]{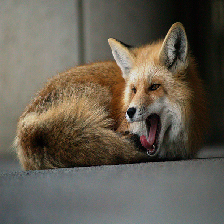}  &
        \includegraphics[width=1.4cm]{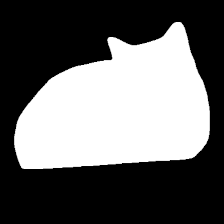} & 
        \includegraphics[width=1.4cm]{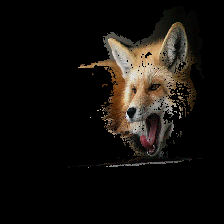}& 
        \includegraphics[width=1.4cm]{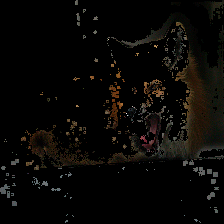} & 
        \includegraphics[width=1.4cm]{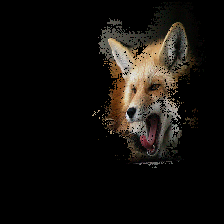}& 
        \includegraphics[width=1.4cm]{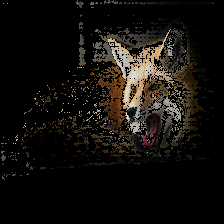} \\
        \hline
        \multicolumn{2}{|c|}{Average distances} & 0.0118 & 0.0584 & 0.0203 & 0.0219 \\
        
        \hline
        \end{tabular}
        
  \caption{Mask and images with 20$\%$ of most relevant pixels using R-LRP and LRP-ab0505 for image 1006 (category 277)}
    \label{fig:1006_pix}
\end{figure}

The average distance between relevant pixels and the object mask measures the dispersion of pixels in the image, and thus indicates whether relevant pixels are grouped or scattered. To measure the average distance, we have created for each object's mask image a distance matrix, where each pixel (matrix element) contains its distance from the mask. Using this matrix, we have been able to calculate the distance of each relevant pixel and then to evaluate an average distance between each image of contributions and the corresponding object mask image (figure \ref{fig:1006_pix}). In order to evaluate the average of this measure over a category, each distance has been normalized because masks have different sizes one to each other.
 
\begin{figure}[!h]  
  \centering
  \begin{tabular}{c c}
  \includegraphics[width=6cm]{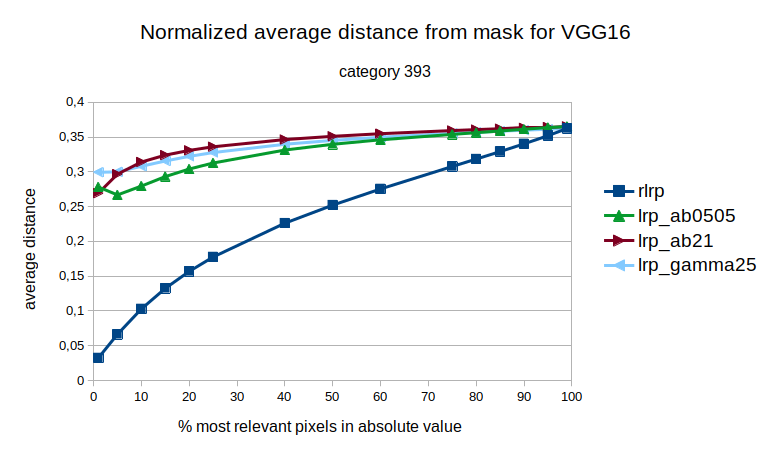}
  &
  \includegraphics[width=6cm]{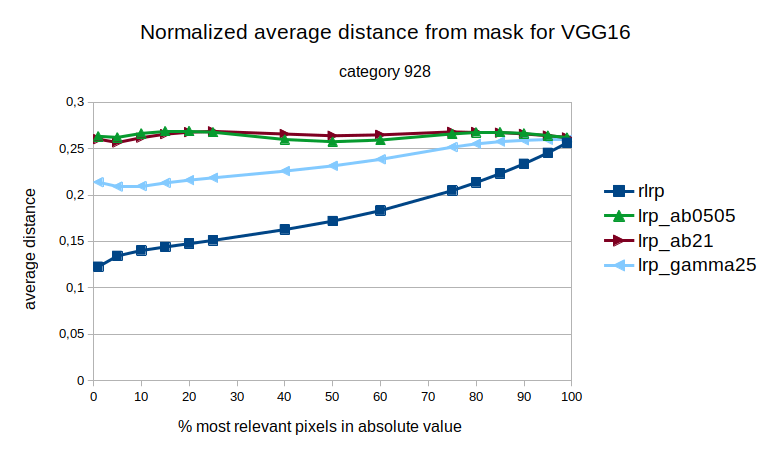}
  \end{tabular}
    \caption{"Best" (cat 393) and "worst" (cat 928) R-LRP average distance for VGG16}
    \label{fig:b_w_distance_vgg16}
\end{figure}

In figure \ref{fig:b_w_distance_vgg16}, we can see that R-LRP method gives better results than LRP-$\alpha\beta$ and LRP-$\gamma$ for categories 393 and 928 for VGG16 in terms of average distance from mask. In fact, R-LRP gives better results than the other methods for all categories tested with VGG16 and VGG19 as R-LRP relevant pixels are more grouped around the masks. Like for pointing game, LRP methods give poor results with category 928 (figure \ref{fig:b_w_distance_vgg16}).

\begin{figure}[!h]  
  \centering
  \begin{tabular}{c c}
  \includegraphics[width=6cm]{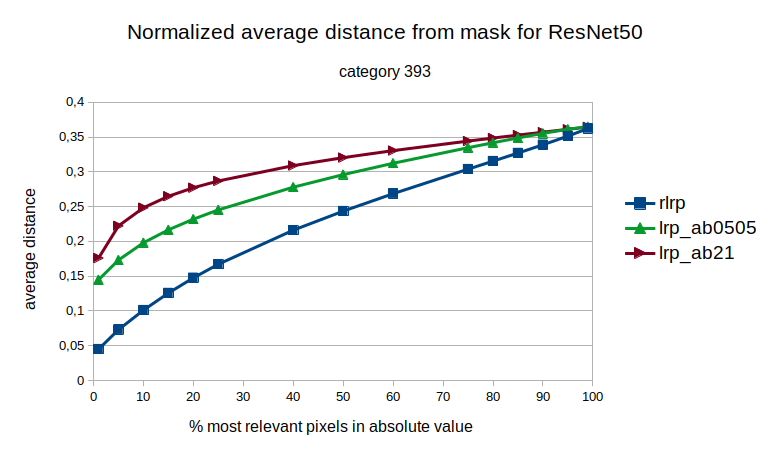}
  &
  \includegraphics[width=6cm]{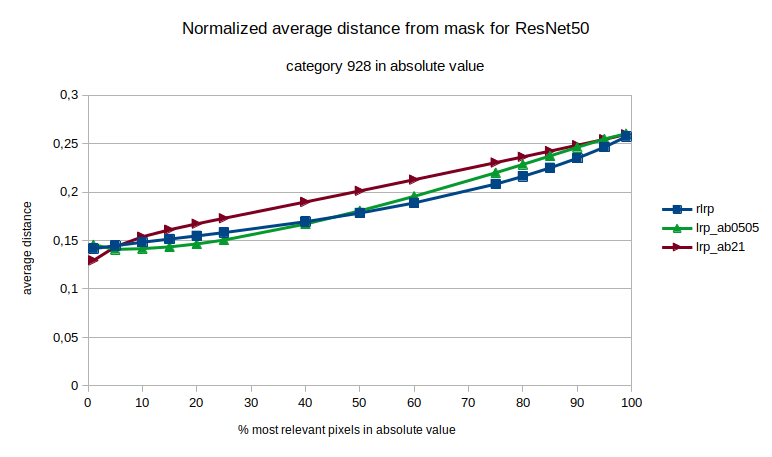}
  \end{tabular}
    \caption{"Best" (cat 393) and "worst" (cat 928) R-LRP average distance for ResNet50}
    \label{fig:b_w_distance_resnet}
\end{figure}

For ResNet50, we give similar results (see figure \ref{fig:b_w_distance_resnet}).

\subsection{Quality evaluation discussion}
\begin{figure}[!h]  
  \centering
       \begin{tabular}{|>{\centering\arraybackslash}m{1.8cm}|>{\centering\arraybackslash}m{1.8cm}|>{\centering\arraybackslash}m{1.8cm}|>{\centering\arraybackslash}m{2.1cm}|>{\centering\arraybackslash}m{1.8cm}|>{\centering\arraybackslash}m{2.1cm}|}
        \hline
        Image& Mask & R-LRP & LRP\_ab0505 & R-LRP & LRP\_ab0505 \\
        & & VGG16 & VGG16 & ResNet50 & ResNet50 \\
        \hline
        {\includegraphics[width=1.4cm]{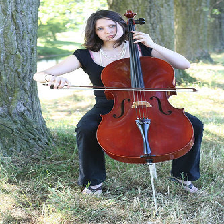} } &
        \includegraphics[width=1.4cm]{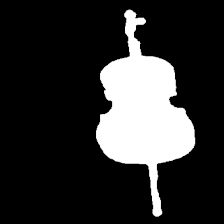} & 
        \includegraphics[width=1.4cm]{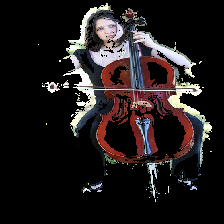}& 
        \includegraphics[width=1.4cm]{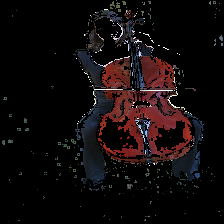} & \includegraphics[width=1.4cm]{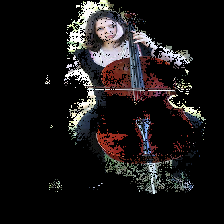}& \includegraphics[width=1.4cm]{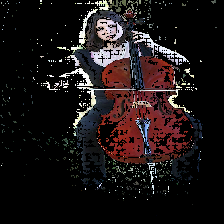} \\ 
        
        \hline
        \end{tabular}
        
  \caption{Mask and images with 20\% of most relevant pixels using R-LRP and LRP-ab0505 for Image 21098 (category 486)}
    \label{fig:21098_pix}
\end{figure}

A quality measurement is a human view because it is based on human criteria such as the identification of an object just use the object itself. So, quality masks for all images in all categories are not sufficient if network categorization takes into account the background in images (see figure \ref{fig:21098_pix}).

\begin{figure}[!h]  
  \centering
       \begin{tabular}{|>{\centering\arraybackslash}m{1.8cm}|>{\centering\arraybackslash}m{1.8cm}|>{\centering\arraybackslash}m{1.8cm}|>{\centering\arraybackslash}m{2.1cm}|>{\centering\arraybackslash}m{1.8cm}|>{\centering\arraybackslash}m{2.1cm}|}
        \hline
        Image& Mask & R-LRP & LRP\_ab0505 & R-LRP & LRP\_ab0505 \\
        & & VGG16 & VGG16 & ResNet50 & ResNet50 \\
        \hline
        {\includegraphics[width=1.4cm]{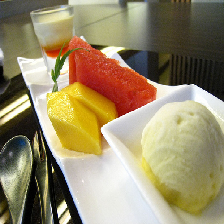} } &
        \includegraphics[width=1.4cm]{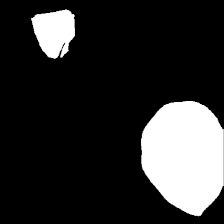} & 
        \includegraphics[width=1.4cm]{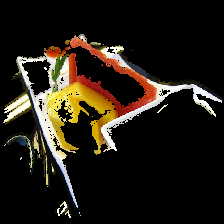}& 
        \includegraphics[width=1.4cm]{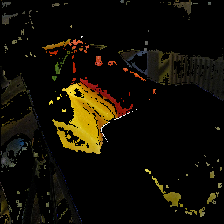} & \includegraphics[width=1.4cm]{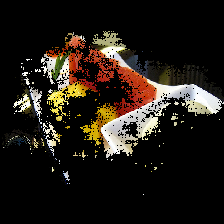}& \includegraphics[width=1.4cm]{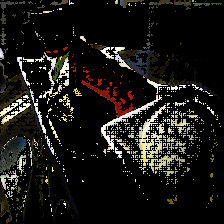} \\ 
        
        \hline
        \end{tabular}
        
  \caption{Mask and images with 20\% of most relevant pixels using R-LRP and LRP-ab0505 for Image 1245 (category 928)}
    \label{fig:1245_pix}
\end{figure}

The relevant pixels have been evaluated with their distances from a mask, but the decision of the network can be done from another criterion. For example, in figure \ref{fig:1245_pix}, for the  image 1245 in category 928, the mask is focus on ice cream objects, but the decision is made with other objects.
 Therefore, average distance from mask is not always a good measure to estimate the relevance of pixels.

Thus, as it was indicated previously, quality evaluations provide just indicative elements on the pixels defined as relevant by a method, with a human view. A method cannot be evaluated just in terms of quality, but quality evaluation can be an additional indicator associated with quantitative ones.

Our R-LRP method outperforms other LRP methods in terms of quantity as, for a small percentage of most relevant pixels, accuracy of images is better for VGG network and also for ResNet50 one (see figures \ref{fig:14} and \ref{fig:resnet50_accu}). It seems that in terms of quality, R-LRP gives the best results for VGG16 (resp. VGG19) and good results for ResNet50.

\section{Conclusion}
\label{conclusion}

Using LRP methods, we can define contributions of inputs onto each output in order to know what a neural network is paying attention to and thus to explain the obtained decision with an image.

In this article, we propose a LRP-method based on a new relative contribution definition denoted R-LRP. The advantage of our method is that, contrary to the previous LRP-methods, no division by values close to zero are needed. Moreover, we don't have any hyperparameters to set or tune and no additional methods are needed.

We compared this method with the previous ones in terms of quality and quantity. The quality is evaluated from the highest pixel/input contributions focused on each object of the class. With our R-LRP method, the contributions seem to be more focused on the object of the class, especially near the boundaries with less spreading pixels around. Moreover, quantity measures (accuracy) achieved on datasets for the same percentage of most relevant inputs/pixels contributions show that, for small percentage of the most relevant pixel contributions, our R-LRP gives better results. Tests have been done keeping different percentage of the most relevant inputs/pixels in images and setting the others to zero.

We have also shown that inputs/pixels with negative contributions are also important in the final decision of the network and make it possible to better localize the object of this decision in the input image.
 
In the future, we will apply R-LRP method to MLP-Mixer \cite{Tolstikhin2021} or Attention \cite{Vaswani2017} architectures.

\acks{
This works was also supported by the Arts \& Métiers Fundation.
}

\nocite{*} 
\printbibliography

@article{Arras2017,
  author =       "Leila Arras and Franziska Horn and Grégoire Montavon and Klaus-Robert Müller and Wojciech Samek",
  title =        "What is relevant in a text document?: An interpretable machine learning approach",
  journal =      "",
  volume =       "",
  number =       "",
  pages =        "",
  year =         "2017",
  DOI =          "10.1371/journal.pone.0181142"
}

@article{Bach2015,
  author =       "Sebastian Bach and Alexander Binder and Grégoire Montavon and Frederick Klauschen and Klaus-Robert Müller and Wojciech Samek",
  title =        "On pixel-wise explanations for non-linear classifier decisions by layer-wise relevance propagation",
  journal =      "PLoS ONE",
  volume =       "",
  number =       "",
  pages =        "",
  year =         "2015",
  DOI =          "10.1371/journal.pone.0130140"
}

@article{DeVore2022,
  author =       "Ingrid Daubechies and Ron DeVore and Simon Foucart and Boris Hanin and Guergana Petrova",
  title =        "Nonlinear Approximation and (Deep) ReLU Networks",
  journal =      "Contructive Approximation",
  volume =       "55",
  number =       "",
  pages =        "127-172",
  year =         "2022",
  DOI =          "10.1007/s00365-021-09548-z"
}

@article{DeVore2021,
  author =       "Ronald DeVore and Boris Hanin and Guergana Petrova",
  title =        "Neural Network Approximation",
  journal =      "Acta Numerica",
  volume =       "30",
  number =       "",
  pages =        "337-444",
  year =         "2021",
  DOI =          "10.1017/S0962492921000052"
}

@article{Horst2019,
  author =       "Fabian Horst and Sebastian Lapuschkin and Wojciech Samek and Klaus-Robert Müller and Wolfgang I. Schöllhorn",
  title =        "Explaining the unique nature of individual gait patterns with deep learning",
  journal =      "Scientific Reports",
  volume =       "9",
  number =       "2391",
  pages =        "",
  year =         "2019",
  DOI =          "10.1038/s41598-019-38748-8"
}

@article{Kohlbrenner2020,
  author =       "Maximilian Kohlbrenner and Alexander Bauer and Shinichi Nakajima and Alexander Binde and Wojciech Samek and Sebastian Lapuschkin",
  title =        "Towards Best Practice in Explaining Neural Network Decisions with LRP",
  journal =      "International Joint Conference on Neural Networks (IJCNN)",
  volume =       "",
  number =       "",
  pages =        "1-7",
  year =         "2020",
  DOI =          "10.1109/IJCNN48605.2020.9206975"
}

@article{Lapuschkin2017,
  author =       "Sebastian Lapuschkin and Alexander Binde and Klaus-Robert Müller and Wojciech Samek",
  title =        "Understanding and Comparing Deep Neural Networks for Age and Gender Classification",
  journal =      "IEEE International Conference on Computer Vision Workshops (ICCVW)",
  volume =       "",
  number =       "",
  pages =        "1629-1638",
  year =         "2017",
  DOI =          "10.1109/ICCVW.2017.191"
}

@article{Lapuschkin2019,
  author =       "Sebastian Lapuschkin and Stephen Wäldchen and Alexander Binde and Grégoire Montavon and Samek Wojciech and Klaus-Robert Müller",
  title =        "Unmasking Clever Hans predictors and assessing what machines really learn",
  journal =      "Nature Communications",
  volume =       "10",
  number =       "1096",
  pages =        "",
  year =         "2019",
  DOI =          "10.1038/s41467-019-08987-4"
}

@article{Jee2021,
  author =       "Yeon-Jee Jung and Seung-Ho Han and Ho-Jin Choi",
  title =        "Explaining CNN and RNN Using Selective Layer-Wise Relevance Propagation",
  journal =      "IEEE Access",
  volume =       "9",
  number =       "",
  pages =        "18670-18681",
  year =         "2021",
  DOI =          "10.1109/ACCESS.2021.3051171"
}

@article{Montavon2017,
  author =       "Grégoire Montavon and Sebastian Lapuschkin and Alexander Binder and Wojciech Samek and Klaus-Robert Müller",
  title =        "Explaining nonlinear classification decisions with deep Taylor decomposition",
  journal =      "Pattern Recognition",
  volume =       "65",
  number =       "",
  pages =        "211-222",
  year =         "2017",
  DOI =          "10.1016/j.patcog.2016.11.008"
}

@article{Montavon2019,
  author =       "Grégoire Montavon  and Alexander Binder and Sebastian Lapuschkin and Wojciech Samek and Klaus-Robert Müller",
  title =        "Layer-Wise Relevance Propagation: An Overview",
  journal =      "Lecture Notes in Computer Science book series",
  volume =       "11700",
  number =       "",
  pages =        "",
  year =         "2019",
  DOI =          ""
}

@article{Fergus2014,
  author =       "Matthew D Zeiler and Rob Fergus",
  title =        "Visualizing and Understanding Convolutional Networks",
  journal =      "Computer Vision – ECCV 2014. ECCV 2014",
  volume =       "8689",
  number =       "",
  pages =        "",
  year =         "2014",
  DOI =          "10.1007/978-3-319-10590-1_53"
}

@article{Petsiuk2018,
  author =       "Vitali Petsiuk and Abir Das and Kate Saenko",
  title =        "RISE: Randomized Input Sampling for Explanation of Black-box Models",
  journal =      "British Machine Vision Conference (BMCV)",
  volume =       "",
  number =       "",
  pages =        "",
  year =         "2018",
  DOI =          "10.48550/arXiv.1806.07421"
}

@article{Dumoulin2016,
  author =       "Vincent Dumoulin and Francesco Visin",
  title =        "A guide to convolution arithmetic for deep learning",
  journal =      "",
  volume =       "",
  number =       "",
  pages =        "",
  year =         "2016",
  DOI =          "10.48550/arXiv.1603.07285"
}

@article{He2016,
  author =       "Kaiming He and Xiangyu Zhang and Shaoqing Ren and Jian Sun",
  title =        "Deep Residual Learning for Image Recognition",
  journal =      "IEEE Conference on Computer Vision and Pattern Recognition",
  volume =       "",
  number =       "",
  pages =        "770-778",
  year =         "2016",
  DOI =          "10.1109/CVPR.2016.90"
}

@Inbook{Holzinger2022,
author="Holzinger, Andreas
and Saranti, Anna
and Molnar, Christoph
and Biecek, Przemyslaw
and Samek, Wojciech",
editor="Holzinger, Andreas
and Goebel, Randy
and Fong, Ruth
and Moon, Taesup
and M{\"u}ller, Klaus-Robert
and Samek, Wojciech",
title="Explainable AI Methods - A Brief Overview",
bookTitle="xxAI - Beyond Explainable AI: International Workshop, Held in Conjunction with ICML 2020, July 18, 2020, Vienna, Austria, Revised and Extended Papers",
year="2022",
publisher="Springer International Publishing",
address="Cham",
pages="13--38",
isbn="978-3-031-04083-2",
doi="10.1007/978-3-031-04083-2_2",
url="https://doi.org/10.1007/978-3-031-04083-2_2"
}

@article{Chung2014,
  author =       "Junyoung Chung and Caglar Gulcehre and KyungHyun Cho and Yoshua Bengio",
  title  =       "Empirical Evaluation of Gated Recurrent Neural Networks on Sequence Modeling",
  journal =      "28th Conference on Neural Information Processing Systems (NIPS) Deep Learning and Representation Learning Workshop",
  volume =       "",
  number =       "",
  pages =        "",
  year =         "2014",
  DOI =          "10.48550/arXiv.1412.3555"
}

@article{Hu2021,
  author =       "Xinyi Huang and Suphanut Jamonnak and Ye Zhao and Tsung Heng Wu and Wei Xu",
  title  =       "A Visual Designer of Layer-wise Relevance Propagation Models",
  journal =      "Computer Graphics Forum",
  volume =       "40",
  number =       "3",
  pages =        "227-238",
  year =         "2021",
  DOI =          "10.1111/cgf.14302"
}

@article{Zhang2018,
  author =       "Jianming Zhang and Sarah Adel Bargal and Zhe Lin and Xiaohui Shen and Jonathan Brandt and Stan Sclaroff",
  title  =       "Top-down Neural Attention by Excitation Backprop",
  journal =      "International Journal of Computer Vision",
  volume =       "126",
  number =       "",
  pages =        "1084-1102",
  year =         "2018",
  DOI =          "10.1007/s11263-017-1059-x"
}

@article{Ribeiro2016,
  author =       "Marco Tulio Ribeiro and Sameer Singh and Carlos Guestrin",
  title  =       "Why Should I Trust You?: Explaining the Predictions of any Classifier",
  journal =      "Proceedings of the 22nd ACM SIGKDD International Conference on Knowledge Discovery and Data Mining",
  volume =       "",
  number =       "",
  pages =        "1135–1144",
  year =         "2016",
  DOI =          "10.1145/2939672.2939778"
}

@article{Hochreiter1997,
  author =       "Sepp Hochreiter and Jürgen Schmidhuber",
  title  =       "Long Short-term Memory",
  journal =      "Neural Computation",
  volume =       "9",
  number =       "8",
  pages =        "1735–1780",
  year =         "1997",
  DOI =          "10.1162/neco.1997.9.8.1735"
}

@article{Tolstikhin2021,
  author =       "Ilya Tolstikhin and Neil Houlsby and Alexander Kolesnikov and Lucas Beyer and Xiaohua Zhai and Thomas Unterthiner and Jessica Yung and Andreas Steiner and Daniel Keysers and Jakob Uszkoreit and Mario Lucic and Alexey Dosovitskiy ",
  title  =       "MLP-Mixer: An all-MLP Architecture for Vision",
  journal =      "35th Conference on Neural Information Processing Systems",
  volume =       "",
  number =       "",
  pages =        "",
  year =         "2021",
  DOI =          "10.48550/arXiv.2105.01601"
}

@article{Covert2021,
  author =       "Ian C. Covert and Scott Lundberg and Su-In Lee",
  title  =       "Explaining by Removing: A Unified Framework for Model Explanation",
  journal =      "Journal of Machine Learning Research",
  volume =       "22",
  number =       "",
  pages =        "1-90",
  year =         "2021",
  DOI =          "10.48550/arXiv.2011.14878"
}

@article{Thomas2019,
  author =       "Armin W. Thomas and Hauke R. Heekeren and Klaus-Robert Müller and Wojciech Samek",
  title =        "Analyzing Neuroimaging Data Through Recurrent Deep Learning Models",
  journal =      "Frontiers in Neuroscience",
  volume =       "13",
  number =       "",
  pages =        "",
  year =         "2019",
  DOI =          "10.3389/fnins.2019.01321"
}

@article{Sundararajan2017,
  author =       "Mukund Sundararajan and Ankur Taly and Qiqi Yan",
  title =        "Axiomatic Attribution for Deep Networks",
  journal =      "ICML'17: Proceedings of the 34th International Conference on Machine Learning",
  volume =       "70",
  number =       "",
  pages =        "3319-3328",
  year =         "2017",
  DOI =          "10.48550/arXiv.1703.01365"
}

@article{Srinivasan2017,
  author =       "Vignesh Srinivasan and Sebastian Lapuschkin and Cornelius Hellge and Klaus-Robert Müller and Wojciech Samek",
  title =        "Interpretable human action recognition in compressed domain",
  journal =      "IEEE International Conference on Acoustics, Speech and Signal Processing (ICASSP)",
  volume =       "",
  number =       "",
  pages =        "1692-1696",
  year =         "2017",
  DOI =          "10.1109/ICASSP.2017.7952445"
}

@article{Selvaraju2017,
  author =       "Ramprasaath R. Selvaraju and Michael Cogswell and Abhishek Das and Ramakrishna Vedantam and Devi Parikh and Dhruv Batra",
  title =        "Grad-CAM: Visual Explanations from Deep Networks via Gradient-based Localization",
  journal =      "IEEE International Conference on Computer Vision (ICCV)",
  volume =       "",
  number =       "",
  pages =        "618-626",
  year =         "2017",
  DOI =          "10.1109/ICCV.2017.74"
}

@article{Simonyan2015,
  author =       "Karen Simonyan and Andrew Zisserman",
  title =        "Very Deep Convolutional Networks for Large-Scale Image Recognition",
  journal =      "3rd International Conference on Learning Representations (ICLR)",
  volume =       "",
  number =       "",
  pages =        "",
  year =         "2015",
  DOI =          "10.48550/arXiv.1409.1556"
}

@article{Springenberg2015,
  author =       "Jost Tobias Springenberg and Alexey Dosovitskiy and Thomas Brox and Martin A. Riedmiller",
  title =        "Striving for Simplicity: The All Convolutional Net",
  journal =      "3rd International Conference on Learning Representations (ICLR)",
  volume =       "",
  number =       "",
  pages =        "",
  year =         "2015",
  DOI =          "10.48550/arXiv.1412.6806"
}

@article{Vaswani2017,
  author =       "Ashish Vaswani and Noam Shazeer and Niki Parmar and Jakob Uszkoreit and Llion Jones and Aidan N. Gomez and Łukasz Kaiser and Illia Polosukhin",
  title  =       "Attention is All you Need",
  journal =      "30th Conference on Neural Information Processing Systems (NIPS)",
  volume =       "",
  number =       "",
  pages =        "",
  year =         "2017",
  DOI =          "10.48550/arXiv.1706.03762"
}

@article{Simao2019a,
  author =       "Miguel Simão and Olivier Gibaru and Pedro Neto",
  title  =       "Online Recognition of Incomplete Gesture Data to Interface Collaborative Robots",
  journal =      "IEEE Transactions on Industrial Electronics",
  volume =       "66",
  number =       "12",
  pages =        "9372-9382",
  year =         "2019",
  DOI =          "10.1109/TIE.2019.2891449"
}

@article{Simao2019b,
  author =       "Miguel Simão and Pedro Neto and Olivier Gibaru",
  title  =       "Improving novelty detection with generative adversarial networks on hand gesture data",
  journal =      "Neurocomputing",
  volume =       "358",
  number =       "",
  pages =        "437-445",
  year =         "2019",
  DOI =          "10.1016/j.neucom.2019.05.064"
}

@article{Simao2019c,
  author =       "Miguel Simão and Pedro Neto and Olivier Gibaru",
  title  =       "EMG-based online classification of gestures with recurrent neural networks",
  journal =      "Pattern Recognition Letters",
  volume =       "128",
  number =       "",
  pages =        "45-51",
  year =         "2019",
  DOI =          "10.1016/j.patrec.2019.07.021"
}

@article{Simao2019d,
  author =       "Miguel Simão and Nuno Mendes and Olivier Gibaru and Pedro Neto",
  title  =       "A Review on Electromyography Decoding and Pattern Recognition for Human-Machine Interaction",
  journal =      "IEEE Access",
  volume =       "7",
  number =       "",
  pages =        "39564-39582",
  year =         "2019",
  DOI =          "10.1109/ACCESS.2019.2906584"
}

@article{Simao2017,
  author =       "Miguel Simão and Pedro Neto and Olivier Gibaru",
  title  =       "Using data dimensionality reduction for recognition of incomplete dynamic gestures",
  journal =      "Pattern Recognition Letters",
  volume =       "99",
  number =       "",
  pages =        "32-38",
  year =         "2017",
  DOI =          "10.1016/j.patrec.2017.01.003"
}

@article{Guerin2021,
  author =       "Joris Guérin and Stephane Thiery and Eric Nyiri and Olivier Gibaru and Byron Boots",
  title  =       "Combining pretrained CNN feature extractors to enhance clustering of complex natural images",
  journal =      "Neurocomputing",
  volume =       "423",
  number =       "",
  pages =        "551-571",
  year =         "2021",
  DOI =          "10.1016/j.neucom.2019.05.064"
}

@article{Guerin2018a,
  author =       "Joris Guérin and Stephane Thiery  and Eric Nyiri and Olivier Gibaru",
  title  =       "Unsupervised robotic sorting: Towards autonomous decision making robots",
  journal =      "International Journal of Artificial Intelligence and Applications (IJAIA)",
  volume =       "9",
  number =       "2",
  pages =        "81-98",
  year =         "2018",
  DOI =          "10.5121/ijaia.2018.9207"
}

@article{Guerin2018b,
  author =       "Joris Guérin and Olivier Gibaru and Eric Nyiri and Stephane Thiery and Byron Boots",
  title  =       "Semantically Meaningful View Selection",
  journal =      "IEEE International Conference on Intelligent Robots and Systems (IROS)",
  volume =       "",
  number =       "",
  pages =        "1061-1066",
  year =         "2018",
  DOI =          "10.1109/IROS.2018.8593524"
}

@book{Graphs92,
    author    = "K. Thulasiraman and M. N. S. Swamy",
    title     = "Graphs: Theory and Algorithms",
    year      = "1992",
    publisher = "John Wiley \& Sons",
    ISBN      = "978-0-471-51356-8",
    DOI=        "10.1002/9781118033104"
}
\end{document}